\documentclass[letterpaper]{article}
\usepackage{graphicx,subfigure}
\usepackage{cite}
\usepackage{ifpdf}
\usepackage{amsmath,amssymb,amsfonts}
\usepackage{textcomp}
\usepackage{algorithmic,algorithm}
\usepackage[margin=1in]{geometry}
\usepackage{multirow}
\usepackage{tabularx}
\usepackage{bbm}
\usepackage{hyperref}

\title{InfoSSM: Interpretable Unsupervised Learning of Nonparametric State-Space Model for Multi-modal Dynamics}
\author{Young-Jin Park, and Han-Lim Choi} 

\begin{document}
\maketitle
\begin{abstract}
The goal of system identification is to learn about underlying physics dynamics behind the time-series data.
To model the probabilistic and nonparametric dynamics model, Gaussian process (GP) have been widely used; GP can estimate the uncertainty of prediction and avoid over-fitting.
Traditional GPSSMs, however, are based on Gaussian transition model, thus often have difficulty in describing a more complex transition model, e.g. aircraft motions. 
To resolve the challenge, this paper proposes a framework using multiple GP transition models which is capable of describing multi-modal dynamics.
Furthermore, we extend the model to the information-theoretic framework, the so-called InfoSSM, by introducing a mutual information regularizer helping the model to learn interpretable and distinguishable multiple dynamics models.
Two illustrative numerical experiments in simple Dubins vehicle and high-fidelity flight simulator are presented to demonstrate the performance and interpretability of the proposed model.
Finally, this paper introduces a framework using InfoSSM with Bayesian filtering for air traffic control tracking.
\end{abstract}

\section{Introduction}
State-space model (SSM) is one of the most general representation model that has been used on a wide range of fields (e.g. aerospace engineering, robotics, economics, and biomedical engineering, etc) for time series analysis \cite{hamilton1994state}. 
The key idea in the state-space model is to construct the latent state space and its transition/observation model representing the sequentially observed data.
In traditional researches and applications, linear Gaussian state-space models are commonly used to solve the state estimation (i.e. inference) or the system identification (i.e. model learning) problems based on Kalman filtering (KF) algorithms \cite{kalman1960new}.
For the last few decades, a huge number of studies have tried to extend to nonlinear SSM such as particle filter \cite{gordon1993novel}, expectation maximization (EM) \cite{briegel1999fisher}, unscented KF \cite{wan2000unscented}, and dual extended kalman filter \cite{wenzel2006dual}, etc.
But, most of the traditional SSM approaches assume a parametric latent model, thus they are only able to be used when we have fairly much information about the dynamics.

Alternatively, nonparametric nonlinear SSMs have been getting attention.
Some recent approaches have design dynamics model and approximate inference structures based on neural networks, and could successfully represent the complex sequential data \cite{karl2016deep, krishnan2017structured, fraccaro2017disentangled, ha2018adaptive}.
On the other hand, the probabilistic dynamics model are known to be more suitable in control problems because they can fully quantify the model uncertainty and enable safe and unbiased model learning \cite{deisenroth2011pilco}.
As such, the probabilistic SSM approaches based on Gaussian process (GP) \cite{rasmussen2006gaussian}, the so-called Gaussian process state-space model (GPSSM), have been widely used in the system identification problems \cite{frigola2014variational, sternberg2017identification, doerr2018probabilistic}.
Although GPSSMs are powerful representation model, they are not still suitable to express a multi-modal dynamics model without any additional probabilistic structure since they assume Gaussian transition model.
For example, if we want to learn the dynamics of airplane motions, single GP doesn't seem to fully represent multiple behaviors (constant velocity, level turn, etc) at once.
In such cases, it is more desirable to assume the latent dynamics model is comprised of multiple processes.
However, GPSSM using multiple GPs has not been reported in the literature yet.
The first key contribution of this paper is the extension of GPSSM framework to multi-modal state-space models, named Hybrid-GPSSM (H-GPSSM).

Meanwhile, GPSSMs are in a class of unsupervised learning that is possibly ill-posed.
Although many unsupervised learning approaches believe their model will be automatically trained as human observer conceives, there is no guarantee they will learn an understandable representation without any regularization because nonparametric models often possess too strong representation power.
In particular, H-GPSSMs have difficulty to identify distinguishable multiple models without human supervision; the data does not include information regarding the types of dynamic models used.
As such, it is never trivial to induce each GP dynamics to learn each mode for multi-modal dynamics.
To resolve similar problems, several recent works in representation learning \cite{chen2016infogan} and reinforcement learning \cite{hausman2017multi, florensa2017stochastic, eysenbach2018diversity} have proposed methods maximizing the mutual information between latent variables and the corresponding output.
However, usage of mutual information regularization in the context of the probabilistic representation learning of time-series data is not yet reported in the literature.
The second key contribution of this paper is the extension of GPSSM to the \textit{interpretable} SSM structure by introducing mutual information regularization between the observation trajectory and the dynamics mode, where the resultant formulation is termed in this paper as InfoSSM.

Note that there have been earlier researches that utilized multiple GPs, called GP experts, to improve the regression performance \cite{rasmussen2002infinite, nguyen2014fast, yuan2009variational}.
Those approaches are based on the idea to divide the input space into subsets and assign a GP for each subset.
However, such approaches can not be used directly into the GPSSM framework since the input space (i.e. latent state space) is unknown in unsupervised learning tasks.
Thereby, this paper proposes a novel approach to assign data into multiple GPs by dividing the output space, not the input space, in an explainable way.

Rest of the paper is organized as follows.
In section \ref{sec:backgrounds}, we begin with a brief summary of the backgrounds about GP and GPSSM.
Section \ref{sec:hgpssm} provides details about our algorithm; we show the mathematical formulation of H-GPSSM and the inference structures to approximate analytically intractable marginal likelihood.
Section \ref{sec:infossm} provides details of InfoSSM, and describes the Bayesian filtering framework that estimates the latent state of the system.
Finally in section \ref{sec:experiments}, we analyze the performance of the proposed model with two system identification experiments and air traffic control (ATC) tracking simulation.

\section{Backgrounds} \label{sec:backgrounds}
\subsection{Nomenclature}
In the following derivations, subscript $n$ denotes $n^{th}$ data set, and subscript $t$ denotes data at $t^{th}$ time step. 
For instance, $\mathbf{X}_t = \{\mathbf{x}_{t,n}\}_{n=1}^N$.
Scalar, vector, and matrix values are, represented using lowercase italic, lowercase bold italic, and capital bold italic, respectively.
$\mathcal{MN}(\mathbf{X}; \mathbf{M}, \mathbf{K}, \mathbf{\Sigma}) = \mathcal{N}(\mbox{vec}(\mathbf{X}); \mathbf{M}, \mathbf{\Sigma} \otimes \mathbf{K})$ represents the probability density of matrix $\mathbf{X}$ under the matrix normal distribution with mean matrix $\mathbf{M}$ and two covariance matrices $\mathbf{K}$ and $\mathbf{\Sigma}$, where $\mathcal{N}$, $\mbox{vec}(\mathbf{X})$, and $\otimes$ denotes normal distribution, the vectorization of $\mathbf{X}$, and the Kronecker product, respectively.
$p(\mathbf{X}|\mathbf{Y};\mathbf{Z})$ denotes the probability of random variable $\mathbf{X}$ conditioning on random variable $\mathbf{Y}$ and parameter $\mathbf{Z}$.

\subsection{Gaussian Process}
GP is a flexible and powerful nonparametric Bayesian model to approximate the distribution over functions \cite{rasmussen2006gaussian}.
Consider we are given a data set of $N$ input $\mathbf{x}_n \in \mathbb{R}^P$ and output $\mathbf{y}_n=f(\mathbf{x}_n) \in \mathbb{R}^Q$ pairs, $\mathcal{D} = \{(\mathbf{x}_1,\mathbf{y}_1), \cdots, (\mathbf{x}_N,\mathbf{y}_N)\}$.
The problem GP seeks to solve is to learn the function $f: \mathbb{R}^P \rightarrow \mathbb{R}^Q$ that maps input space to output space of data \footnote{In this paper, we particularly consider matrix-variate GP which further considers the covariance among multi-output.}.
GP assumes function outputs $\mathbf{F} = [\mathbf{y}_1, \cdots, \mathbf{y}_N] \in \mathbb{R}^{N\times Q}$ are jointly Gaussian:
\begin{equation}
p(\mathbf{F};\mathbf{X}) = \mathcal{MN}(\mathbf{F}; m(\mathbf{X}), \mathbf{K}_{\mathbf{X}\mathbf{X}}, \boldsymbol{\Sigma})
\end{equation}
where $\mathbf{X} = [\mathbf{x}_1, \cdots, \mathbf{x}_N] \in \mathbb{R}^{N\times P}$, $m(\cdot)$ is the mean function,  $\mathbf{K}$ is the GP kernel matrix, and $\boldsymbol{\Sigma}$ is the covariance matrix among multi-output.
Affine mean function and squared exponential (SE) automatic relevance determination (ARD) kernel is commonly used for the covariance matrix between $\mathbf{x}$ and $\mathbf{x}'$:
\begin{equation}
m(\mathbf{x}) = \mathbf{H}\mathbf{x} + \mathbf{b} ~~~~~~~~~~~~~~~~
\mathbf{K}_{\mathbf{x}\mathbf{x}'} = \sigma_f^2\exp{(-\frac{1}{2}\sum_p(\frac{x_p-x_p'}{\lambda_p})^2)}
\end{equation}
where subscript $p$ denotes $p^{th}$ dimension of the variable ($\mathbf{x} = \{x_p\}_{p=1}^P$), while $\mathbf{H} \in \mathbb{R}^{Q\times P}$, $\mathbf{b}\in \mathbb{R}^{Q}$, $\sigma_f$, $\lambda_p$ are hyperparameters of mean and covariance function to learn.
Given input points $\mathbf{X}$ and function outputs $\mathbf{F}$, the probability distribution of function value at the new input location $\mathbf{x}^*$ can be predicted as normal distribution:
\begin{equation}
p(\mathbf{f}^*;\mathbf{F}, \mathbf{x}^*, \mathbf{X}) = \mathcal{N}(\mathbf{f}^*; r(\mathbf{x}^*;\mathbf{F},\mathbf{X}), v(\mathbf{x}^*;\mathbf{X})\boldsymbol{\Sigma})
\end{equation}
with mean $r(\mathbf{x}^*;\mathbf{F},\mathbf{X}) = m(\mathbf{x}^*) + \mathbf{k}_{\mathbf{x}^*\mathbf{X}}\mathbf{K}_{\mathbf{X}\mathbf{X}}^{-1}(\mathbf{F} - m(\mathbf{X}))$ and covariance $v(\mathbf{x}^*;\mathbf{X}) = k_{\mathbf{x}^*\mathbf{x}^*} - \mathbf{k}_{\mathbf{x}^*\mathbf{X}}\mathbf{K}_{\mathbf{X}\mathbf{X}}^{-1}\mathbf{k}_{\mathbf{X}\mathbf{x}^*}$.

However, GPs are known to suffer from extensive $\mathcal{O}(N^3)$ computational cost due to the matrix inversion of $N \times N$ covariance matrix, $\mathbf{K}_{\mathbf{X}\mathbf{X}}$.
To cope with the problem, sparse approximation is commonly used \cite{snelson2006sparse, titsias2009variational}, which introduces $M$ inducing inputs $\mathbf{Z} = [\mathbf{z}_1, \cdots, \mathbf{z}_M] \in \mathbb{R}^{M\times P}$ and outputs $\mathbf{U} = [\mathbf{u}_1, \cdots, \mathbf{u}_M] \in \mathbb{R}^{M\times Q}$.
Prior distribution of sparse GP is given by
\begin{equation}
p(\mathbf{U};\mathbf{Z}) = \mathcal{MN}(\mathbf{U}; m(\mathbf{Z}), \mathbf{K}_{\mathbf{Z}\mathbf{Z}}, \boldsymbol{\Sigma}) .
\end{equation}
Assuming inducing variables can sufficiently represent the distribution of original GP function, the true GP prediction is approximated as
\begin{equation}
p(\mathbf{f}^*;\mathbf{F}, \mathbf{x}^*, \mathbf{X}) \approx p(\mathbf{f}^* \mid \mathbf{U};\mathbf{x}^*, \mathbf{Z}) = \mathcal{N}(\mathbf{f}^*; r(\mathbf{x}^* \mid \mathbf{U};\mathbf{Z}), v(\mathbf{x}^*;\mathbf{Z})\boldsymbol{\Sigma}). 
\end{equation}
Note that inducing outputs are random variables to infer whereas inducing inputs are treated as parameters to learn.
Resultant computational complexity is reduced to $\mathcal{O}(NM^2)$.
By an abuse of notation, we drop $ \mid \mathbf{U}$ and $;\mathbf{Z}$ for the GP prediction term in the following derivations.

\subsection{Gaussian Process State-Space Model}
\begin{figure}[t!]
	\centering
	\includegraphics[width=.5\textwidth]{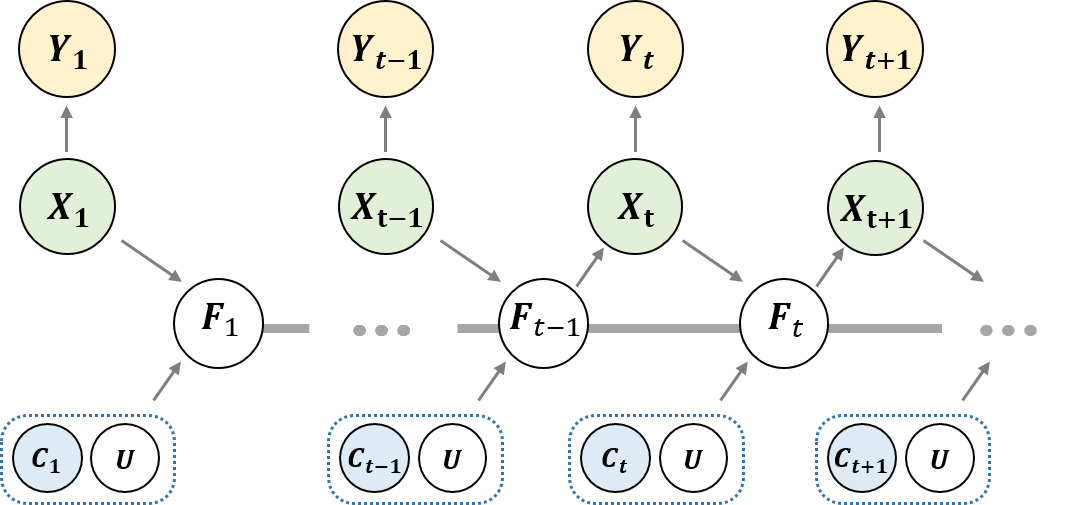}
	\caption{The graphical model of H-GPSSM. H-GPSSM reduces into GPSSM when the dynamics code $\mathbf{C}_t$ is leaved out. }
	\label{InfoSSM}
\end{figure}

State-space model is a representation model to describe the dynamic system, consisted of latent Markovian state $\mathbf{x}_t$ and observation output $\mathbf{y}_t$ at time $t$, which is given by
\begin{align}
\dot{\mathbf{x}}_t &= f(\mathbf{x}_t) + \boldsymbol{\epsilon}_f \nonumber \\
\mathbf{y}_t       &= g(\mathbf{x}_t) + \boldsymbol{\epsilon}_g
\end{align}
where $f$ and $g$ are transition and observation model, while $\boldsymbol{\epsilon}_f$ and $\boldsymbol{\epsilon}_g$ are process and measurement noise. 
$\mathbf{p}_t$, $\mathbf{v}_t$, and $\mathbf{a}_t$ signify the position, velocity, acceleration vector.
In this paper, we focus on the problem where the transition model is completely unknown (i.e. non-parametric) whereas observation model is fairly known (i.e. parametric) \footnote{It is easy to extend GPSSM to the model adopting both non-parametric transition model and observation model. However, such setting brings out the problem of \textit{severe non-identifiability} between $f$ and $g$ \cite{frigola2014variational, doerr2018probabilistic}, and degrade the interpretability of the learned latent model.}.
In the discrete canonical state-space model with the predefined time interval $\delta t$, latent state at time $t+1$ can be recursively approximated as
$\mathbf{x}_{t+1} = \mathbf{x}_{t} +  \delta t \cdot \dot{\mathbf{x}}_t$ .
For the sake of notational simplicity, we denote $\hat{\mathbf{f}}_{t} = f(\mathbf{x}_t)$ and $\mathbf{f}_{t} = \mathbf{x}_{t} +    \delta t \cdot \hat{\mathbf{f}}_t$.
The key concept of GPSSM is modeling transition model of the system as GP function:
\begin{align}
p(\hat{\mathbf{f}}_t  \mid \mathbf{x}_t, \mathbf{U}) &= \mathcal{N}(\hat{\mathbf{f}}_t; r(\mathbf{x}_t), v(\mathbf{x}_t)\boldsymbol{\Sigma}) \nonumber \\
p(\mathbf{f}_{t}  \mid \mathbf{x}_t, \mathbf{U}) &= \mathcal{N}(\mathbf{f}_{t}; \mathbf{x}_t + \delta t \cdot  r(\mathbf{x}_t), \delta t^2 \cdot v(\mathbf{x}_t)\boldsymbol{\Sigma}) .
\end{align}
As shown in the graphical model of GPSSM in Fig. \ref{InfoSSM}, the joint distribution of GPSSM variables for time step $t=1, \cdots, T$ is factorized as
\begin{align}
p(\mathbf{Y}_{1:T}, \mathbf{X}_{1:T}, \mathbf{F}_{1:T-1}, \mathbf{U}) &= p(\mathbf{Y}_{1:T} \mid \mathbf{X}_{1:T})p(\mathbf{X}_{1:T}, \mathbf{F}_{1:T-1} \mid \mathbf{U})p(\mathbf{U}) \nonumber \\
&= \left[ \prod_{t=1}^{T} p(\mathbf{Y}_t \mid \mathbf{X}_t) \right] \left[ p(\mathbf{X}_1) \prod_{t=1}^{T-1} p(\mathbf{X}_{t+1} \mid \mathbf{F}_t) p(\mathbf{F}_t \mid \mathbf{X}_t, \mathbf{U}) \right] p(\mathbf{U}) .
\label{eq:GPSSM}
\end{align}

\section{GPSSM with Multiple Transition Models} \label{sec:hgpssm}
\subsection{Multiple State-Space Modeling}
Previous GPSSM studies \cite{frigola2014variational, sternberg2017identification, doerr2018probabilistic} mostly targeted on the problem where the control input sequence is given.
However, the control input is often unobservable and the only data we can access is the observation outputs, $\mathbf{y}_{1:T}$.
For instance, when we observe maneuvering airplanes, we can receive position (and velocity) data from radar or GPS signal but the corresponding control signals (i.e. thrust, and control surfaces, etc) are not given.
Yet it is impractical to consider every possible control input sequences or infer the unknown control input.
Alternatively, inspired from the interacting multiple model (IMM) algorithm \cite{bar1989tracking}, we propose the model, Hybrid-GPSSM, that approximates the dynamics model as the combination of $L$ numbers of motion patterns:
\begin{equation}
\dot{\mathbf{x}_t} \approx \sum_{l=1}^L \mathbbm{1}(c_t=l)[f_l(\mathbf{x}_t) + \boldsymbol{\epsilon}_{f, l}]\label{eq:multiple}
\end{equation}
where $\mathbbm{1}(\cdot)$ is an indicator function, and $f_l$ and $\boldsymbol{\epsilon}_{f, l}$ are $l^{th}$ transition model and process noise, respectively.
As in the GPSSM structure, each transition model is represented by GP with $p(\mathbf{U}_l) = \mathcal{MN}(\mathbf{U}_l; m(\mathbf{Z}_l), \mathbf{K}_{\mathbf{Z}_l\mathbf{Z}_l}, \boldsymbol{\Sigma}_l)$, $\mathbf{Z} = \{\mathbf{Z}_1, \cdots, \mathbf{Z}_L\}$ and $\mathbf{U} = \{\mathbf{U}_1, \cdots, \mathbf{U}_L\}$:
\begin{equation}
p(\hat{\mathbf{f}}_{t}  \mid  \mathbf{x}_{t}, c_t, \mathbf{U}) = \sum_{l=1}^L p(c_t=l) p(\hat{\mathbf{f}}_t \mid \mathbf{x}_{t}, \mathbf{U}_{l}) = \sum_{l=1}^L p(c_t=j) \mathcal{N}(\hat{\mathbf{f}}_{t}; r_{l}(\mathbf{x}_{t}), v_{l}(\mathbf{x}_{t})\boldsymbol{\Sigma}_{l}) .
\end{equation}
where $r_{l}$ and $\mathbf{v}_{l}$ is mean and variance function of $l^{th}$ GP prediction.
In this paper, we call $c_t$ as a \textit{latent dynamics code} at time $t$ in the way that $c_t$ determines the mode of dynamics governing the transition of latent states from time $t$ to $t+1$.
We model the transition model of the latent dynamics code as a Markov chain with mode transition matrix $\mathbf{P}$:
\begin{equation}
p(c_{t+1} = j | c_t = i) = P_{i,j} \label{eq:jump}
\end{equation}
where $P_{i,j}$ is the $i^{th}$ row and $j^{th}$ column element of the matrix $\mathbf{P}$, and $\sum_{j=1}^L P_{i,j} = 1$ for all $i \in \{1, \cdots, L\}$.
Note that, when the mode transition matrix is the identity matrix, there is no mode transition along the observation sequence.
\footnote{In practice, we could not find an advantage for using trainable transition matrix. Transition matrix converges to identity matrix unless the trajectory is very long.}

The joint distribution variables of H-GPSSM is then given by
\begin{align}
& p(\mathbf{Y}_{1:T}, \mathbf{X}_{1:T}, \mathbf{F}_{1:T-1}, \mathbf{C}_{1:T-1}, \mathbf{U}) = p(\mathbf{Y}_{1:T} \mid \mathbf{X}_{1:T})p(\mathbf{X}_{1:T}, \mathbf{F}_{1:T-1} \mid \mathbf{C}_{1:T-1}, \mathbf{U}) p(\mathbf{C}_{1:T-1}) p(\mathbf{U}) \nonumber \\
&= \left[ \underbrace{\prod_{t=1}^{T} p(\mathbf{Y}_t \mid \mathbf{X}_t)}_{observation} \right]
\left[ \underbrace{p(\mathbf{X}_1) \prod_{t=1}^{T-1} p(\mathbf{X}_{t+1} \mid \mathbf{F}_t) p(\mathbf{F}_t \mid \mathbf{X}_t, \mathbf{C}_t, \mathbf{U})}_{initial ~ states ~ and ~ transition} \right]
\left[ \underbrace{p(\mathbf{C}_1) \prod_{t=1}^{T-2} p(\mathbf{C}_{t+1} \mid \mathbf{C}_t)}_{initial ~ codes ~ and ~ transition} \right]
\left[ \underbrace{\prod_{l=1}^L p(\mathbf{U}_l)}_{GP} \right]. \label{eq:InfoSSM}
\end{align}
But, $\mathbf{C}_t$ unfortunately are hidden latent variables, which make the problem difficult to solve.

\subsection{Inference Sturcture of Hybrid-GPSSM}
The goal of GPSSM is to optimize GP hyperparameters and inducing variables maximizing the marginal likelihood $p(\mathbf{Y}_{1:T})$, so as to learn the latent dynamics that most represents the data.
Unlike single-layer GP, however, GPSSM is no longer GP as it propagates recursively along the time.
This causes a challenge for the estimation of the marginal likelihood that contains intractable integration.
Instead of directly optimizing the marginal likelihood, variational inference approaches alternatively maximize the lower bound of the objective, which is called 
the evidence lower bound (ELBO):
\begin{align}
&\log p(\mathbf{Y}_{1:T}) =  \log \int p(\mathbf{Y}_{1:T} \mid \mathbf{X}_{1:T}, \mathbf{F}_{1:T-1}, \mathbf{C}_{1:T-1}, \mathbf{U})p(\mathbf{X}_{1:T}, \mathbf{F}_{1:T-1}, \mathbf{C}_{1:T-1}, \mathbf{U}) \nonumber \\
&\geq \mathbb{E}_{q(\mathbf{X}_{1:T}, \mathbf{F}_{1:T-1}, \mathbf{C}_{1:T-1}, \mathbf{U})}\left[\log \frac{p(\mathbf{Y}_{1:T} \mid \mathbf{X}_{1:T}, \mathbf{F}_{1:T-1}, \mathbf{C}_{1:T-1}, \mathbf{U})p(\mathbf{X}_{1:T}, \mathbf{F}_{1:T-1}, \mathbf{C}_{1:T-1}, \mathbf{U})}{q(\mathbf{X}_{1:T}, \mathbf{F}_{1:T-1}, \mathbf{C}_{1:T-1}, \mathbf{U})}\right] \equiv \mathcal{L}(\mathbf{\mathbf{Y}}_{1:T}) . \label{eq:elbo}
\end{align}
This paper follows a doubly stochastic variational inference approach  \cite{salimbeni2017doubly, doerr2018probabilistic}, known to give superior performance to other variational approach \cite{damianou2013deep} or EM \cite{bui2016deep} for deep GP structures; it neither impose any assumptions about the independence between GP layers nor Gaussianity for GP outputs.
Similar to \cite{doerr2018probabilistic}, we factorize the variational distribution as
\begin{align}
&q(\mathbf{X}_{1:T}, \mathbf{F}_{1:T-1}, \mathbf{C}_{1:T-1}, \mathbf{U}) 
= q(\mathbf{X}_{1:T-1}, \mathbf{F}_{1:T-1} \mid \mathbf{C}_{1:T-1}, \mathbf{U}) q(\mathbf{C}_{1:T-1}) q(\mathbf{U}) \nonumber \\
&= \left[ q(\mathbf{X}_{1}) \prod_{t=1}^{T-1} p(\mathbf{X}_{t+1} \mid \mathbf{F}_t) p(\mathbf{F}_t \mid \mathbf{X}_t, \mathbf{C}_{1:T-1}, \mathbf{U}) \right]  q(\mathbf{C}_{1:T-1}) q(\mathbf{U}) \label{eq:var} .
\end{align}
Remark that the gap between the log marginal likelihood and the lower bound decreases as the variational distribution $q$ gets closer to the true posterior, hence it is very important to select a proper inference structure.

\begin{figure}[t!]
	\centering
	\includegraphics[width=\textwidth]{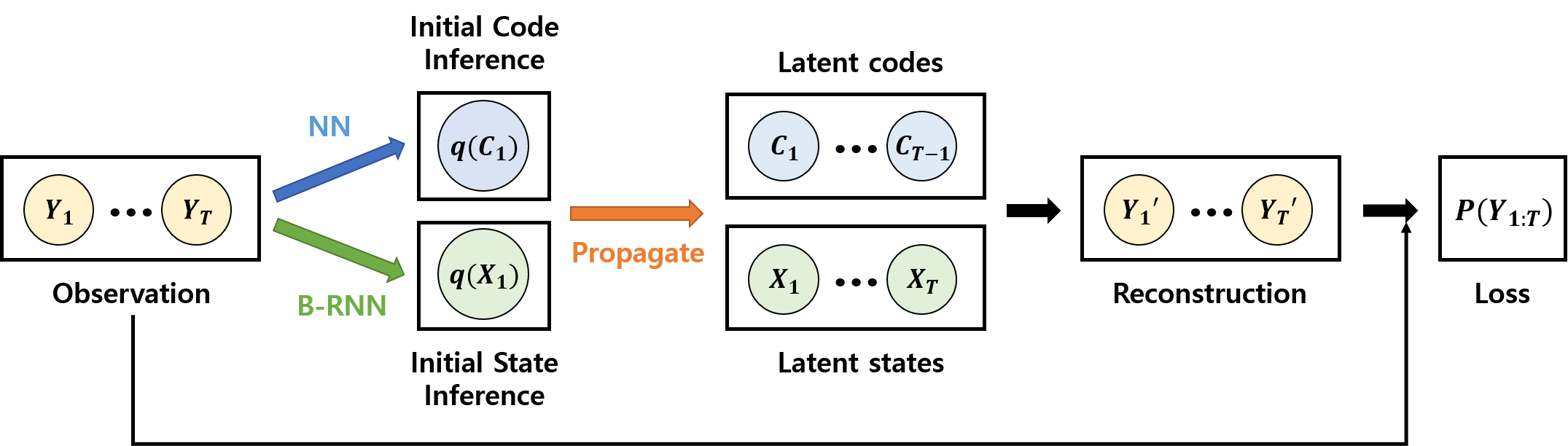}
	\caption{The overall inference structure for Hybrid-GPSSM}
	\label{arch}
\end{figure}

\subsubsection{Approximate Inference for Latent states}
Approximate inference for the posterior distribution of latent states consists of three parts: variational distribution of GP inducing points $q(\mathbf{U})$, initial state $q(\mathbf{X}_1)$, and propagated states $q(\mathbf{X}_{2:T})$. 
Primarily, $q(\mathbf{U})$ is factorized by $L$ numbers of $q(\mathbf{U}_l)$ which is parameterized by a matrix-variate normal distribution sharing the same multi-output covariance with each GP prior:
\begin{align}
q(\mathbf{U}) &= \prod_{l=1}^L q(\mathbf{U}_l) \nonumber \\
q(\mathbf{U}_l) &= \mathcal{MN}(\mathbf{A}_l, \mathbf{S}_l, \boldsymbol{\Sigma}_l)
\end{align}
where $\mathbf{A}_l \in \mathbb{R}^{M \times Q}$ and $\mathbf{S}_l \in \mathbb{R}^{M \times M}$ are variational parameters to optimize.

\begin{figure}[hbt!]
	\centering
	\includegraphics[width=.5\textwidth]{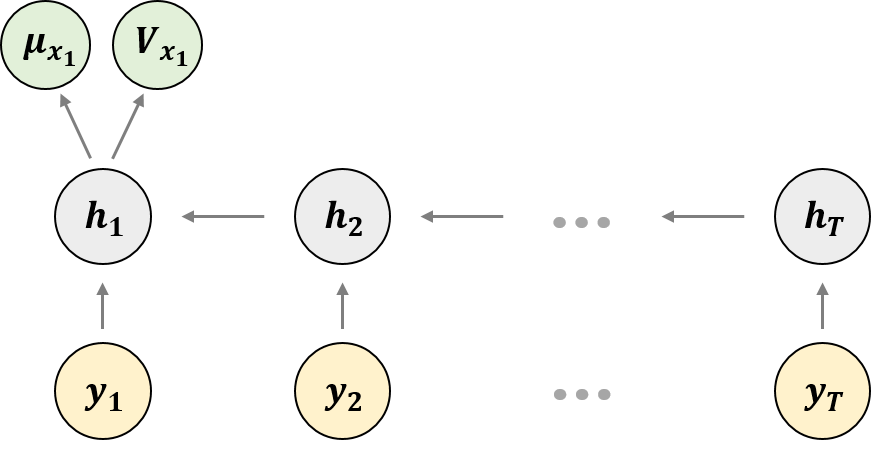}
	\caption{The inference structure for the initial latent state. $\mathbf{h}_t$ is a $t^{th}$ hidden state}
	\label{BRNN}
\end{figure}

Secondly, for the initial state distribution, one naïve approach is to parameterize $q(\mathbf{x}_{1,n})$ per data. Though it may give fairly exact inference results, it is not only computationally expensive but impossible to generalize for unseen data.
Alternatively, this paper uses amortized inference networks \cite{kingma2013auto, rezende2015variational} to approximate the posterior distribution.
We choose backward recurrent neural network (BRNN) that is designed to mimic Kalman smoothing algorithm \cite{krishnan2017structured}; the initial hidden state of BRNN $\mathbf{h}_1$ is expected to contain the compressed information from $\mathbf{y}_{T}, \cdots, \mathbf{y}_{1}$.
An additional shallow neural network outputs the mean and covariance of initial states from $\mathbf{h}_1$:
\begin{align}
q(\mathbf{X}_1) &= \prod_{n=1}^N q(\mathbf{x_{1,n}}) \nonumber \\
q(\mathbf{x}_{1,n}) &= \mathcal{N}(\mathbf{x}_{1,n} ; \boldsymbol{\mu}_\phi(\mathbf{y}_{1:T,n}), \mathbf{V}_\phi(\mathbf{y}_{1:T,n})) \label{eq:init}
\end{align}
where $\phi$ is a set of parameters within a BRNN and a neural-network.
To reduce the computational complexity, diagonal covariance is used.
The inference structure is shown in Fig. \ref{BRNN}.

Finally, propagated states can be recursively derived from GP prediction:
\begin{align}
q(\mathbf{X}_{2:T}) &= \prod_{n=1}^N q(\mathbf{x_{2:T,n}}) \nonumber \\
q(\mathbf{x}_{t+1,n} \mid \mathbf{x}_{t,n}, c_{t,n}) &= p(\mathbf{x}_{t+1,n}|\mathbf{f}_{t,n})q(\mathbf{f}_{t,n} \mid \mathbf{x}_{t,n}, c_{t,n}) \nonumber \\
q(\hat{\mathbf{f}}_{t,n} \mid \mathbf{x}_{t,n}, c_{t,n})
&= \int p(\hat{\mathbf{f}}_{t,n} \mid \mathbf{x}_{t,n}, c_{t,n}, \mathbf{U}) q(\mathbf{U})d\mathbf{U} \nonumber \\
&= \sum_{l=1}^L p(c_{t,n}=l) \cdot \mathcal{N}(\hat{\mathbf{f}}; \tilde{r}_{l}(\mathbf{x}_{t,n}), \tilde{v}_{l}(\mathbf{x}_{t,n})\boldsymbol{\Sigma}_{l}) \label{eq:propagate}
\end{align}
where
$ \tilde{r}(\mathbf{x}) = m(\mathbf{x}) + \mathbf{k}_{xZ}\mathbf{K}_{ZZ}^{-1}(\mathbf{A} - m(\mathbf{Z}))$ 
and
$\tilde{v}(\mathbf{x}) = k_{xx} - \mathbf{k}_{xZ}\mathbf{K}_{ZZ}^{-1}(\mathbf{K}_{ZZ}-\mathbf{S})\mathbf{K}_{ZZ}^{-1}\mathbf{k}_{Zx}$ \cite{salimbeni2017doubly, park2018deep}. 

\subsubsection{Approximate Inference for Latent Dynamics Code}
The probability distribution of $\mathbf{c}_{1:T-1,n}$ can be ideally inferred from $\mathbf{y}_{1:T,n}$ although intractable.
But it appears that the mode of dynamics can be distinguished from the shape of the observation trajectory.
Thus, we built a neural network, $Q_\phi$, that receives the observation trajectory as input and outputs the corresponding initial latent dynamics code:
\begin{align}
q(\mathbf{C}_{1:T-1}) &= \prod_{n=1}^N q(\mathbf{c}_{1:T-1,n}) \nonumber \\
q(\mathbf{c}_{1:T-1,n}) &= q(c_{1,n}) \prod_{t=1}^{T-2} p(c_{t+1,n} \mid c_{t,n}) \nonumber \\
q(c_{1,n}) &= Cat(c_{1,n} ; Q_\phi(\mathbf{y}_{1:T, n}))
\label{eq:code}
\end{align}
where $Cat(x; \mathbf{p}=\{p_1, \cdots, p_L\})$ is a categorical distribution with probability mass function: $p(x=l) = p_l$ for $l = 1, \cdots, L$.
A softmax activation is used for the last layer to mimic the distribution.

\subsection{Monte Carlo Objective}
Based on (\ref{eq:InfoSSM} - \ref{eq:var}), the ELBO is derived:
\begin{align}
&\mathcal{L}(\mathbf{\mathbf{Y}}_{1:T})
= \sum_{t=1}^{T} \mathbb{E}_{q(\mathbf{X}_t)} \left[ \log p(\mathbf{Y}_t \mid \mathbf{X}_t) \right]
+ \mathbb{E}_{q(\mathbf{X}_1)}\left[\log \frac{p(\mathbf{X}_{1})}{q(\mathbf{X}_{1})}\right]
+ \mathbb{E}_{q(\mathbf{C}_1)}\left[\log \frac{p(\mathbf{C}_1)}{q(\mathbf{C}_1)}\right]
+ \mathbb{E}_{q(\mathbf{U})}\left[\log \frac{p(\mathbf{U})}{q(\mathbf{U})}\right] \nonumber \\
&= \sum_{n=1}^{N} \Bigg( \sum_{t=1}^{T} \mathbb{E}_{q(\mathbf{x}_{t,n})} \left[ \log p(\mathbf{y}_{t,n} \mid \mathbf{x}_{t,n}) \right]
+ \mathbb{E}_{q(\mathbf{x}_{1,n})}\left[\log \frac{p(\mathbf{x}_{1,n})}{q(\mathbf{x}_{1,n})}\right]
+ \mathbb{E}_{q(c_{1,n})}\left[\log \frac{p(c_{1,n})}{q(c_{1,n})}\right] \Bigg)
- \sum_{l=1}^{L} KL(q(\mathbf{U}_l)||p(\mathbf{U}_l)) .
\end{align}
Expectation terms can be computed by Monte Carlo sampling approach for ELBO update.
Meanwhile, instead of using traditional ELBO update, this paper adopts Monte Carlo objective (MCO) update approach \cite{mnih2016variational} that is known to not only increase the representation power of variational distribution but give a tighter lower bound by using multiple samples in a more effective way \cite{burda2015IWAE, cremer2017reinterpreting}.
The MCO with $K$ Monte Carlo samples is given by
\begin{align}
\mathcal{L}^K(\mathbf{\mathbf{Y}}_{1:T})
&=\sum_{n=1}^{N} \mathbb{E}_{q(\mathbf{x}_{1:T,n}^{(k)}, \mathbf{c}_{1:T-1,n}^{(k)})} \left[ \log \frac{1}{K}\sum_{k=1}^K\Big( \prod_{t=1}^{T}p(\mathbf{y}_{t,n} \mid \mathbf{x}_{t,n}^{(k)}) \cdot \frac{p(\mathbf{x}_{1,n}^{(k)})}{q(\mathbf{x}_{1,n}^{(k)})} \cdot \frac{p(c_{1,n}^{(k)})}{q(c_{1,n}^{(k)})} \Big) \right]
- \sum_{l=1}^{L} KL(q(\mathbf{U}_l)||p(\mathbf{U}_l)) .
\label{eq:MCO}
\end{align}
See Appendix for detailed derivations.
During the computation, reparameterization trick \cite{kingma2013auto, rezende2015variational, jang2016categorical} is further used to make learning signal from the MCO be back-propagated into every inference network.
As such, we construct the end-to-end training procedure in a fully differentiable manner.

\section{InfoSSM} \label{sec:infossm}
\subsection{Mutual Information Regularization for Latent Dynamics Code}
Ideally, we hope each GP is trained in a meaningful way even without any supervision. 
For example, when we have a data set of trajectories from the maneuvering airplane, we expect one GP learns constant velocity, another learns level turn, and the others learn climb and descend motions.
But in the worst case, it is also possible that only one useful GP dynamics is trained to represent every motion while the others learn useless motions \cite{bacon2017option}.
To avoid such circumstance, we need to introduce an additional information theoretic objective that lead the latent dynamics code to satisfies two criterions.
First, the code should contain \textit{useful} information determining the shape of trajectory. 
Second, trajectories generated from different GPs should be \textit{distinguishable}.
As such, there should be the strong mutual dependence between the latent dynamics code sequence and the corresponding trajectory.

Inspired from \textit{InfoGAN} \cite{chen2016infogan}, this paper as well adopts mutual information regularization for latent dynamics code.
In the information theoretic sense, the mutual information is a measure quantifying the mutual dependence between two random variables:
\begin{equation}
\mathcal{I}(V; X) = H(V) - H(V \mid X) = H(X) - H(X \mid V) .
\end{equation}
Mutual information can also be interpreted as the amount of information contained in one random variable about the other random variable.
If two random variables are mutually independent, mutual information becomes zero.
Finally, the InfoSSM is trained to maximize the mutual information along with the MCO:
\begin{equation}
\mathcal{L}_{Info}^K(\mathbf{\mathbf{Y}}_{1:T}) 
= \mathcal{L}^K(\mathbf{\mathbf{Y}}_{1:T}) + \lambda \mathcal{I}(\mathbf{c}_{1:T-1}; \mathcal{G}(x_1, \mathbf{c}_{1:T-1}))
\end{equation}
where $\lambda$ is a tunable parameter adjusting the relative scale between MCO and regularization.
$\mathcal{G}$ is a function sampling the observation trajectory from the learned H-GPSSM model for given initial state and code sequence by using \eqref{eq:multiple} and \eqref{eq:propagate}. 
Note that, however, it is intractable to directly compute the mutual information term.
Instead, we optimize the lower bound of the mutual information by variational information maximization approach \cite{barber2003algorithm}:
\begin{align}
\mathcal{I}(\mathbf{c}_{1:T-1}; \mathcal{G}(x_1, \mathbf{c}_{1:T-1})) 
&= H(\mathbf{c}_{1:T-1}) - H(\mathbf{c}_{1:T-1} \mid \mathcal{G}(x_1, \mathbf{c}_{1:T-1})) \nonumber \\
&\ge H(\mathbf{c}_{1:T-1}) + \mathbb{E}_{\mathbf{c}_{1:T-1} \sim p(\mathbf{c}_{1:T-1}), ~ \mathbf{y}_{1:T} \sim \mathcal{G}(x_1, \mathbf{c}_{1:T-1})} \big[ \log q(\mathbf{c}_{1:T-1} ; \mathbf{y}_{1:T}) \big] \nonumber \\
&\approx H(c_1) + \mathbb{E}_{\mathbf{c}_{1:T-1} \sim p(\mathbf{c}_{1:T-1}), ~ \mathbf{y}_{1:T} \sim \mathcal{G}(x_1, \mathbf{c}_{1:T-1})} \big[ \log q(c_1 ; \mathbf{y}_{1:T}) \big].
\label{eq:MI}
\end{align}
See Appendix for detailed derivations. 
$H(c_1)$ is constant since the prior is fixed, thus we did not compute the term in the implementation.
Unlike previous studies \cite{chen2016infogan, hausman2017multi, florensa2017stochastic, eysenbach2018diversity}, we did not need to build a new auxiliary function $q$ since we already have the approximate inference model $Q_\phi$ in \eqref{eq:code}.
Expectation term can be computed by Monte Carlo estimation, and detail implementation is provided in Algorithm \ref{al:InfoSSM}. 
We implement InfoSSM with Tensorflow~\cite{abadi2016tensorflow}, and Adaptive moment estimation (Adam) algorithm \cite{kingma2014adam} is used for gradient-descent optimizer.
The code is soon available at \href{https://github.com/yjparkLiCS/InfoSSM}{\textit{https://github.com/yjparkLiCS/InfoSSM}}.

\subsection{Approximate Bayesian Filtering and Smoothing Using InfoSSM} \label{sec:Filter}
By integrating the inference structure of InfoSSM with Bayesian filtering technique, the more accurate inferences of latent state trajectory become possible.
Instead of simply propagating with the transition model as in the Eq.\eqref{eq:propagate}, it is desirable to recursively estimate the latent state for each time step by using the observation data.
A Commonly used Kalman Filter, however, is not suitable for InfoSSM which contains non-linear transition models with non-Gaussian noise.
Alternatively, particle filter (PF), one of the most well known non-linear data assimilation methods, is selected in this work. 
\footnote{Since Gaussian processes are analytically differentiable, InfoSSM can be also combined with the IMM-EKF filtering algorithm.}

A PF is sample-based Bayes filter, where the state probability distribution is represented as:
\begin{equation}
p(\mathbf{x}) \approx \sum_{k=1}^{K} w^{(k)} \delta(\mathbf{x} - \mathbf{x}^{(k)} )
\end{equation}
where $\mathbf{x}^{(k)}$ and $w^{(k)}$ is the state and the weight for the $k^{th}$ of $K$ particles, respectively.
For each time $t$, the particle is propagated via the (learned) transition model while the weight of each particle of state $\mathbf{x}_t^{(k)}$ is given as its likelihood under given observation $\mathbf{y}_t^{(k)}$:
\begin{equation}
w_t^{(k)} = w_t^{(k-1)} \cdot p(\mathbf{x}_t^{(k)} \mid \mathbf{y}_t) . \label{eq:weight}
\end{equation}

After the propagation step, resampling step is applied when the effective sample size (ESS) drops below than the certain number (e.g. $K/2$), to resolve the sample degeneracy problem.
The weight of each particle is reset to $1/K$ after resampling step.
Note that the resampling step can be implemented in a differentiable way, thus it can be applied not only after the training but also during the training stage as introduced in \cite{maddison2017filtering}.
Empirically found that, however, powerful inference through resampling step prevents the InfoSSM to learn accurate dynamics models.
Thus, we used particle filtering only for the target tracking.
A detail implementation of target tracking framework using InfoSSM is provided in Algorithm \ref{al:Filter} and\ref{al:PF}.

\begin{algorithm} [t!]
	\caption{InfoSSM Implementation}
	\label{al:InfoSSM}
	\begin{algorithmic}[1]
		\STATE Initialize $\lambda$.
		\FOR {$iter = 1$ to $max\_iter$}
		\STATE $\mathcal{L}_{Info}^K \leftarrow 0$.
		\FOR {$n = 1$ to $N$}
		\STATE Select $n^{th}$ observation data: $\mathbf{y}_{1:T,n}$ from $\mathcal{D}$.
		\STATE Sample $K$ initial states and code sequences: $\{(\mathbf{x}_{1,n}^{(k)},\mathbf{c}_{1:T-1,n}^{(k)})\}_{k=1}^{K}$, using \eqref{eq:init} and \eqref{eq:code}.
		\STATE Propagate states: $\{\mathbf{x}_{1:T,n}^{(k)}\}_{k=1}^{K}$, using \eqref{eq:propagate} recursively.
		\STATE Compute the MCO: $\mathcal{L}^K(\mathbf{y}_{1:T,n})$, using \eqref{eq:MCO}.
		\STATE $\mathcal{L}_{Info}^K \leftarrow \mathcal{L}_{Info}^K + \mathcal{L}^K(\mathbf{y}_{1:T,n})$.
		\ENDFOR
		\STATE Sample $\mathbf{c}_{1:T-1} \sim p(\mathbf{c}_{1:T-1})$.
		\STATE Generate the trajectories $\mathbf{y}_{1:T}$ from each $\mathbf{c}_{1:T-1}$.
		\STATE Compute the \textbf{mutual information}: $\mathcal{I}$, using \eqref{eq:MI}.
		\STATE $\mathcal{L}_{Info}^K \leftarrow \mathcal{L}_{Info}^K + \lambda \mathcal{I}$.
		\STATE Update InfoSSM using Adam.
		\ENDFOR
	\end{algorithmic} z\textsl{z}
\end{algorithm}

\begin{algorithm}[t!]
	\caption{Target Tracking Using InfoSSM}
	\label{al:Filter}
	\begin{algorithmic}[1]
		\REQUIRE $\mathbf{y}_{t-T+1:t}$: past observation sequence, $K$: the number of particles
		\STATE Sample $K$ initial states and code: $\{(\mathbf{x}_{t-T+1}^{(k)},\mathbf{c}_{t-T+1}^{(k)})\}_{k=1}^{K}$. ~~~~ // Smoothing
		\STATE Initialize weights, $\{w^{(k)}_{t-T+1}\}$, as $1/K$.
		\FOR {$\tau = t-T+2$ to $t$ and $k = 1$ to $K$}
		\STATE Propagate $\{\mathbf{x}^{(k)}_{\tau}, \mathbf{c}^{(k)}_{\tau}, w^{(k)}_{\tau}\} $ from $\{\mathbf{x}^{(k)}_{\tau-1}, \mathbf{c}^{(k)}_{\tau-1}, w^{(k)}_{\tau-1}\}$, using PF (Algorithm \ref{al:PF}). ~~~~ // Filtering
		\ENDFOR
		\STATE Estimate $p(\mathbf{x}_{t}) \approx \sum_{k=1}^{K} w_{t}^{(k)} \delta(\mathbf{x}_{t} - \mathbf{x}_{t}^{(k)} )$.
		~~~~ // Estimation
	\end{algorithmic}
\end{algorithm}

\section{Experiments} \label{sec:experiments}
\subsection{Dubins Vehicle}
First, we evaluate the proposed algorithm with the Dubins vehicle experiment.
The Dubins path is commonly used approximated dynamics in the planning and control problems for wheeled vehicles and airplanes \cite{lavalle2006planning, tsourdos2010cooperative, beard2012small}.
The dynamics of Dubins' vehicle is given by:
\begin{align}
\dot{\mathbf{p}_x} &= V \cos \theta \nonumber \\
\dot{\mathbf{p}_y} &= V \sin \theta \nonumber \\
\dot{\theta} &= u + \epsilon_{f,true} .
\end{align}
where $V$ is the speed of vehicle which is assumed as constant ($V=1$), and $\epsilon_{f,true} \sim \mathcal{N}(0, 0.1)$.
Depending on the control input, Dubins' vehicle shows three motion primitives: right(R), straight(S), left(L) with $u = -1, 0, 1$, respectively.
Fifty sub-trajectories are used for the training procedure with the observation noise $\mathcal{N}(0, 0.1)$ for each dimension.
In this toy example, we intentionally cut sub-trajectories in a way that there is no control input change in the middle of the sub-trajectory and fix the transition matrix as identity matrix in order to make an ideal experiment environment for the InfoSSM.

\subsection{High-Fidelity Flight Simulation}
To demonstrate the performance of our model for the practical situation, we evaluate the proposed algorithm with flight trajectory data generated from X-plane 11 simulator.
The type of airplane was Cessna 172 skyhawk, and it flew in San Diego, California, U.S.
One hundred sub-trajectories are used for the training procedure with the observation noise $\mathcal{N}(0, 5)$ meters for each dimension.
In this example, we considered the mode transition unlike the Dubins example, so let the transition matrix as a trainable variable initialized by the identity matrix.

\subsection{Anlaysis of Learned Models}
As baselines, we compare the InfoSSM with the state art of the art GPSSM model, PRSSM \cite{doerr2018probabilistic} ($L = 1$ and $\lambda = 0$), and InfoSSM without mutual information regularization ($\lambda = 0$ ) which is equivalent to H-GPSSM.
We evaluate the models in three aspects: interpretability, model accuracy, and long-term prediction performance.
We would refer the readers to Appendix C for experiment details.

\begin{figure}[t!]
	\centering
	\subfigure[Dubins: H-GPSSM ($\lambda = 0$)]{
		\includegraphics[width=.45\columnwidth]{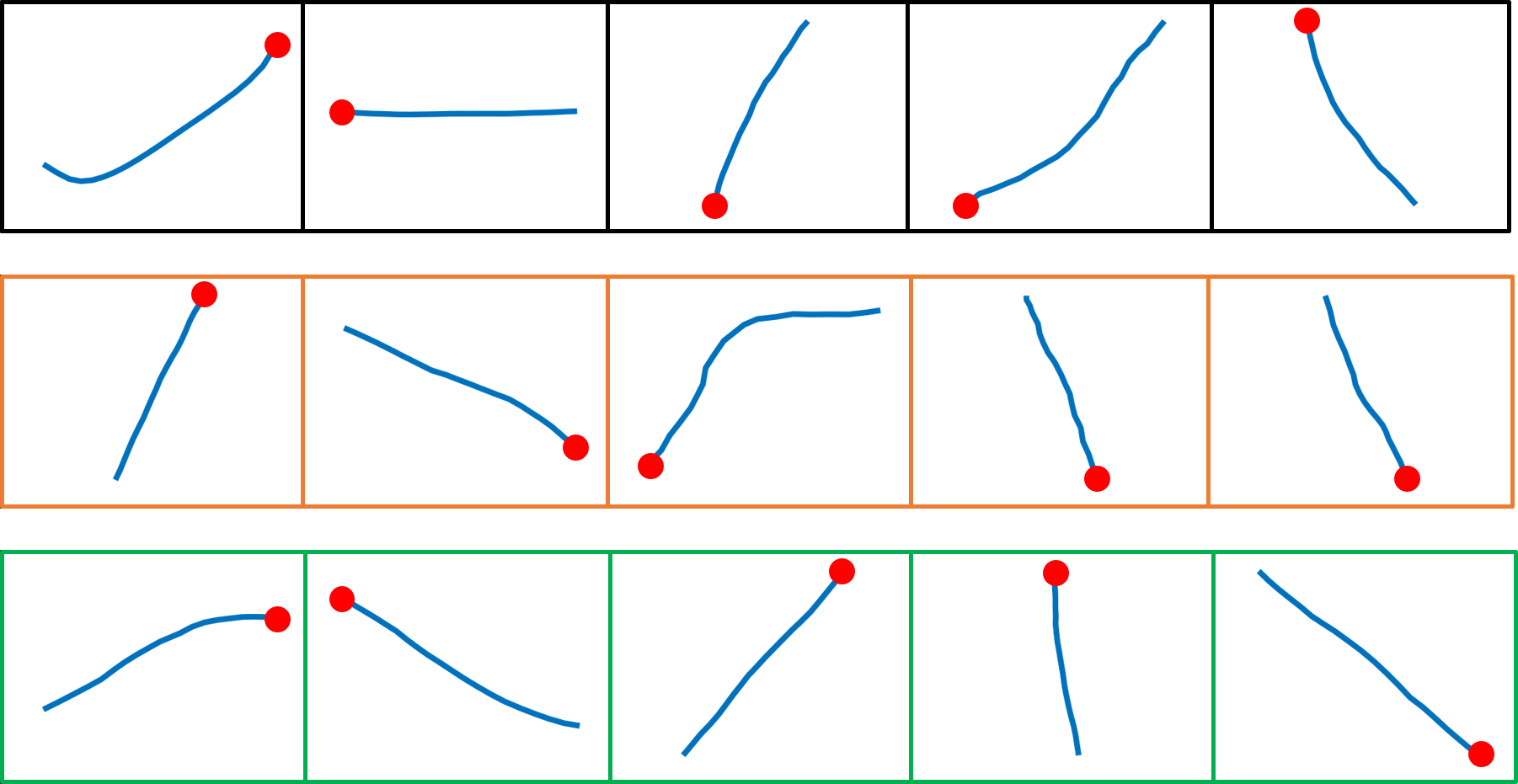}}
	\subfigure[Dubins: InfoSSM ($\lambda > 0$)]{
		\includegraphics[width=.45\columnwidth]{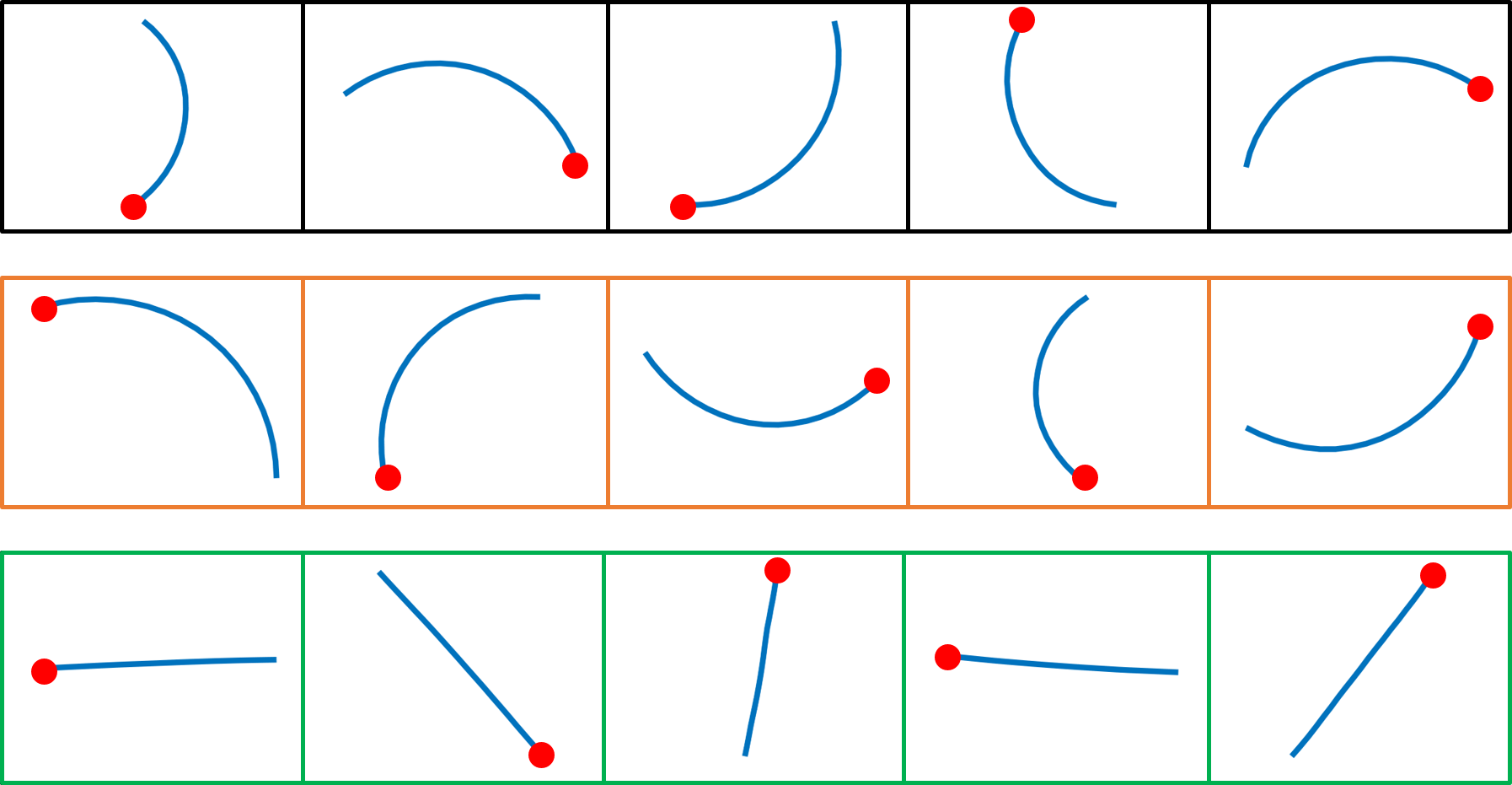}}
	\caption{The sampled Dubins trajectories from different codes. Initial state is marked with red dot. From above, $\mathbf{c}$ = 1, 2, 3. As shown in results, the latent dynamics code for H-GPSSM does not seem to play a significant role.}
	\label{fig:interp}
\end{figure}

\subsubsection{Interpretability}
\begin{figure}[h!]
	\centering
	\subfigure[Cessna: H-GPSSM ($\lambda = 0$)]{
		\includegraphics[width=.45\columnwidth]{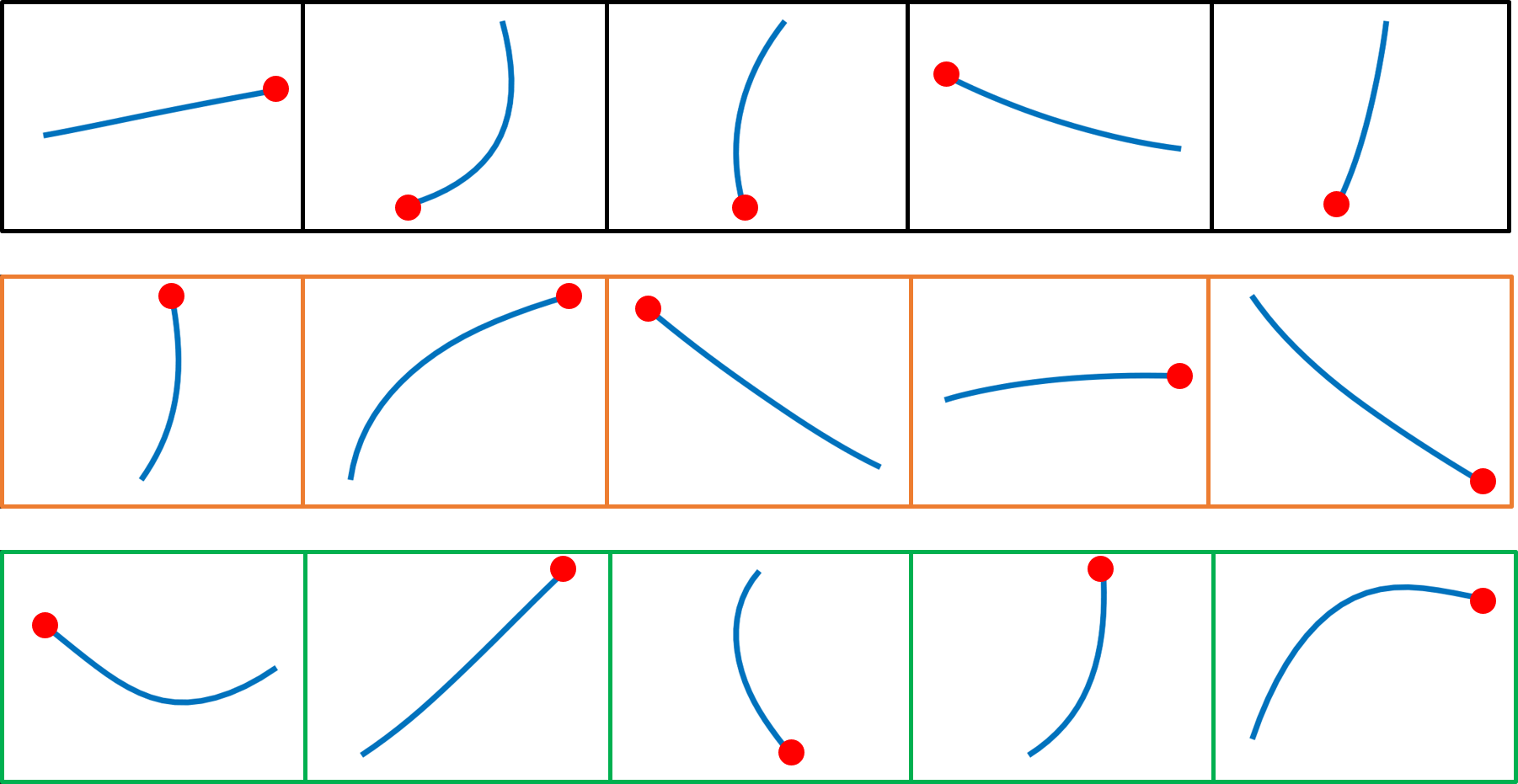}}
	\subfigure[Cessna: InfoSSM ($\lambda > 0$)]{
		\includegraphics[width=.45\columnwidth]{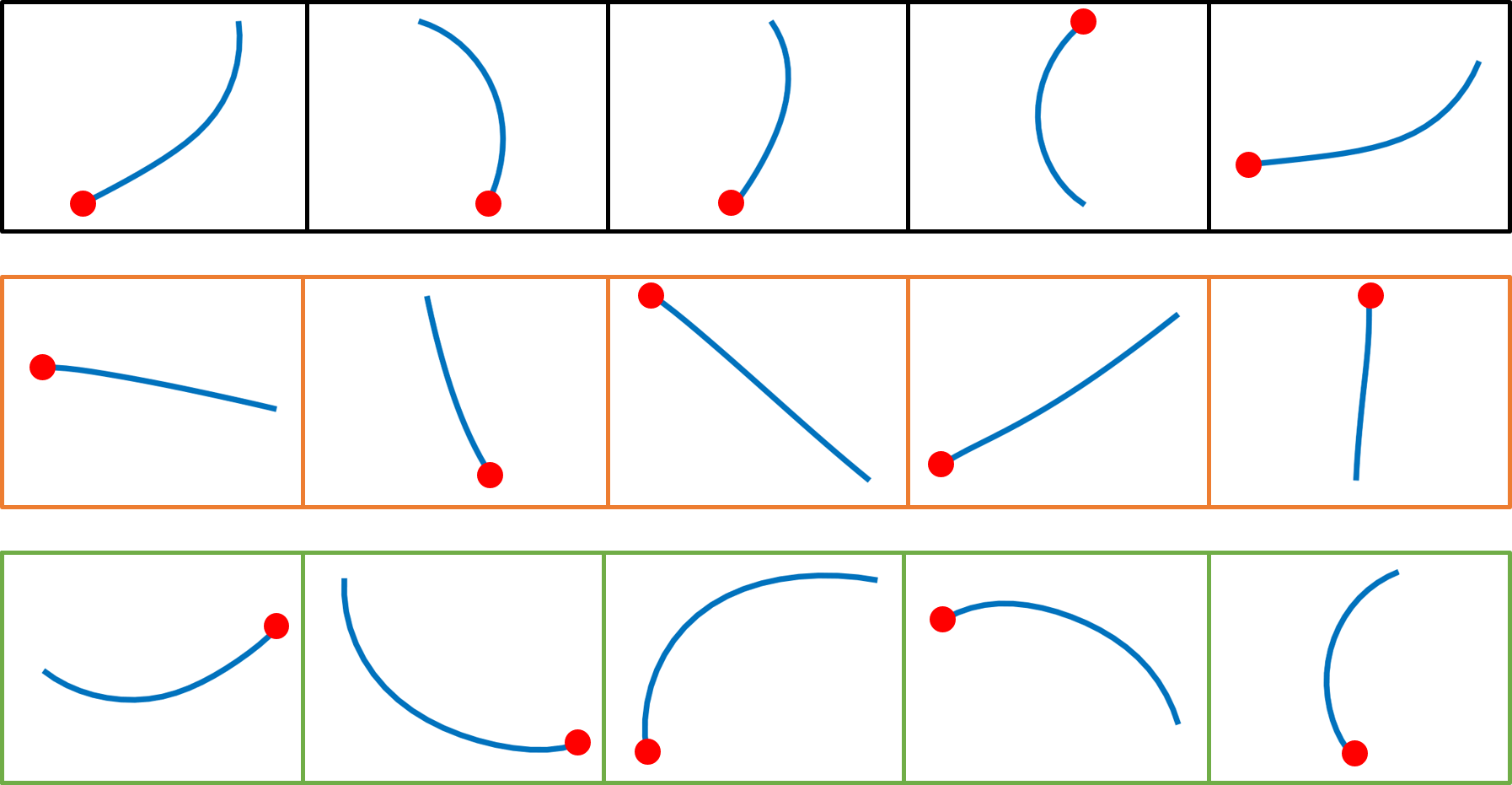}}
	\caption{The sampled Cessna airplane trajectories from different codes. Initial state is marked with red dot. From above, $\mathbf{c}$ = 1, 2, 3.}
	\label{fig:interp2}
\end{figure}

Most importantly, we analyze the effect of mutual information to model interpretability.
We compare the results between InfoSSM and H-GPSSM.
To visualize the effect, several trajectories generated from different GP (i.e. latent dynamics code) at random initial states are plotted in Fig. \ref{fig:interp} and Fig. \ref{fig:interp2}.
As shown in the results, we can't see any interpretable meanings from H-GPSSM.
We observe that the model without mutual information regularization tends to utilize only one GP; it tries to learn the motion considering overall transition, which is a very inefficient way.
The InfoSSM, in contrast, successfully learns disentangled representation so that each code perfectly distinguishes L, S, and R motion pattern.
Furthermore, as shown in Table \ref{tab:MCO}, InfoSSM demonstrates the highest mutual information thus the most interpretable.

\subsubsection{Model Accuracy}
To compare the model accuracy (i.e. how well model represents the observation), we analyze two factors: the lower bound of log marginal likelihood and the reconstruction (i.e. inference) performance for 50 Dubins and 100 Cessna test trajectories which are not in training data set.
To investigate the performance of the learned model itself, Bayesian filtering is not used during reconstruction processes.
30 samples are used for the reconstruction.

Primarily, note that the H-GPSSM often show worse performance than PRSSM which uses only single GP.
This is due to the fact that H-GPSSM fails to use multiple GPs while the model complexity is increased and make GPs hard to train.
Thanks to the mutual information regularization, on the other hand, InfoSSM successfully distinguishes data from different dynamics pattern, and gradient signal from each data group could be efficiently delivered to each GP.
As a result, InfoSSM not only achieves the highest lower bound as shown in Table \ref{tab:MCO}, but shows the smallest root mean square error and largest log-likelihood compare to H-GPSSM and PRSSM for both Dubins and Cessna airplane systems.
Further reconstruction results are presented Table \ref{tab:reconC}, \ref{tab:reconD} and Fig. \ref{fig:reconC}, \ref{fig:reconD} in Appendix D.

\begin{table}[t!]
	\caption{\label{tab:MCO} The Model Accuracy}
	\centering
	\begin{tabular}{|c|ccc|ccc|}
		\hline
		& \multicolumn{3}{c|}{Dubins} & \multicolumn{3}{c|}{Cessna} \\ 
		& InfoSSM & H-GPSSM & PRSSM & InfoSSM & H-GPSSM & PRSSM \\ \hline
		$\mathcal{L}^K$ & \textbf{3727.2}  &  1249.2  & 1525.5 & \textbf{-33110.8}  &  -429933.9    & -78155.3\\ 
		$\mathcal{I}$   &  \textbf{-0.0020} & -2.874     & -    &  \textbf{-0.0021} & -1.513     & -   \\
		RMSE (test)   &  \textbf{0.6217} & 2.2507     & 1.8430   &  \textbf{0.9082$\mathbf{\times10^3}$} & 1.512$\times10^3$     & 1.365$\times10^3$   \\ 
		log-likelihood (test)  &  \textbf{40.119} & -3.124  & -14.996 &  \textbf{-419.40} & -699.94     & -608.49\\ \hline  
	\end{tabular}
\end{table}

\subsubsection{Long-term Prediction}
\begin{figure}[hbt!]
	\centering
	\subfigure[InfoSSM ($\lambda > 0$)]{
		\includegraphics[width=0.65\columnwidth]{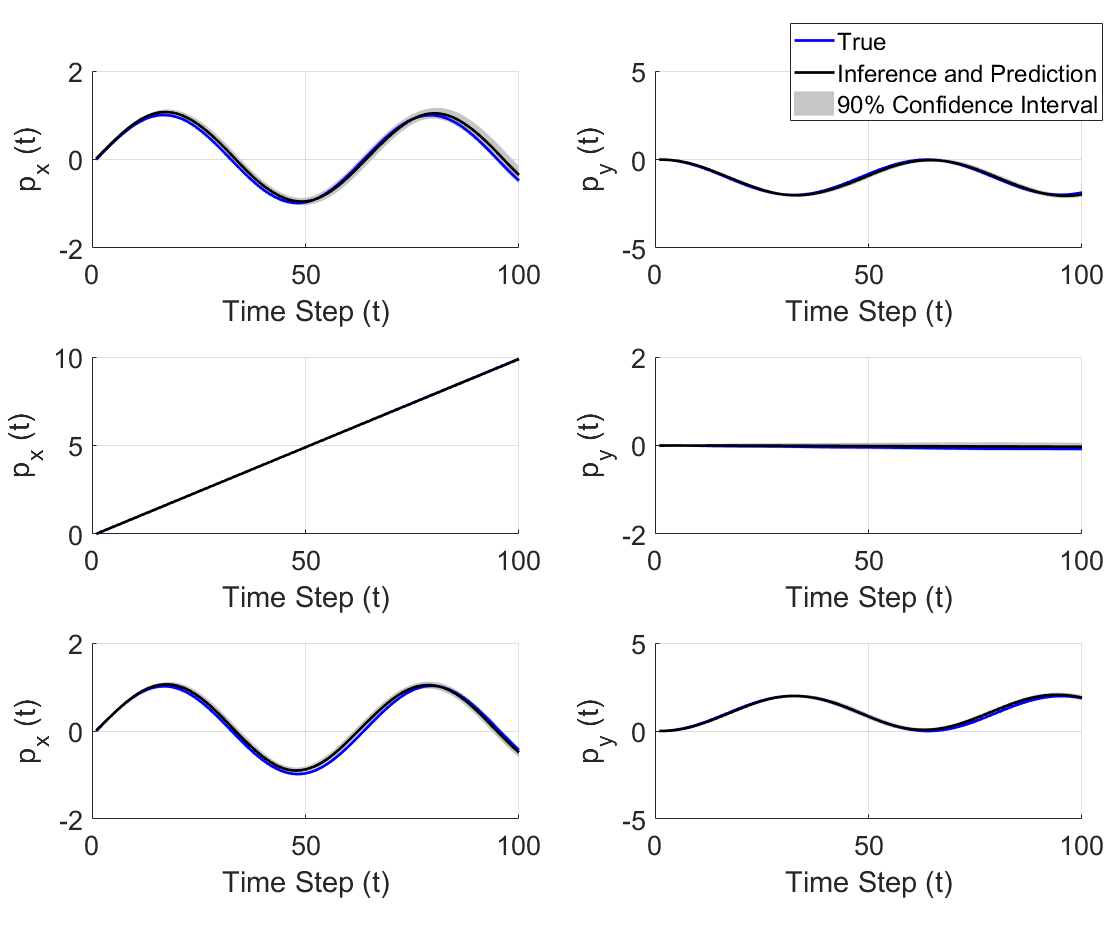}}
	\subfigure[H-GPSSM ($\lambda = 0$)]{
		\includegraphics[width=0.45\columnwidth]{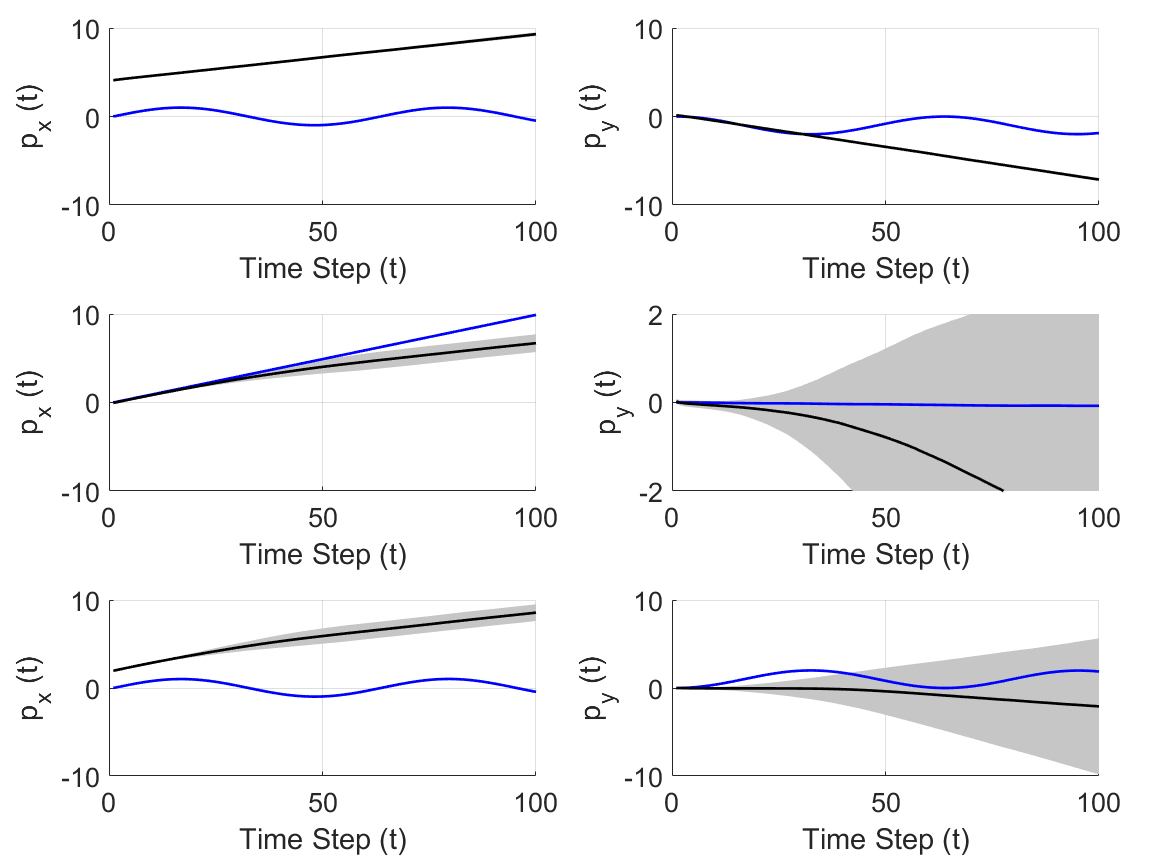}}
	\subfigure[PRSSM]{
		\includegraphics[width=0.45\columnwidth]{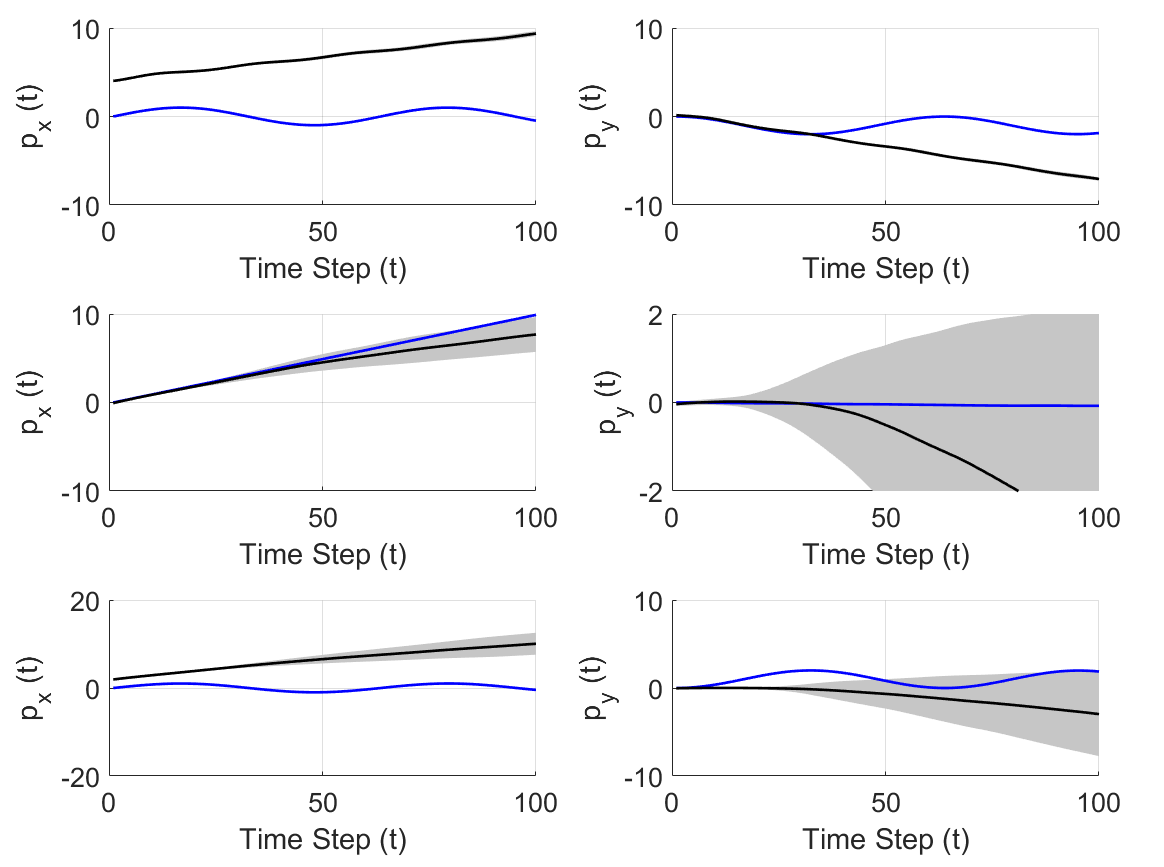}}
	\caption{The long term prediction for Dubins dynamics.}
	\label{fig:predict}
\end{figure}

Finally, we evaluate the long-term prediction performance of InfoSSM to see whether the learned dynamics well matches with the true dynamics.
From the initial state $\mathbf{x}_1 = [0,0,1,0]$, we propagate the Dubins vehicle for 100 time steps with $u = -1, 0, 1$. \footnote{For Cessna airplane, we could not manipulate ground truth trajectories with constant motion pattern since the true dynamics is unknown, unlike the Dubins model.}
Using the first 20 step trajectory, the model inferred the code and latent state.
Then, the latent state is propagated with the learned dynamics of corresponding code and compared with the true trajectory.
As shown in Fig. \ref{fig:predict}, the InfoSSM shows highly accurate long-term prediction for each dynamics mode.
Note that, however, baseline models fail to predict the future state and the uncertainty is increased rapidly as time goes.

\begin{table}[t!]
	\caption{\label{tab:ATC} The Tracking Accuracy}
	\centering
	\begin{tabular}{|c|ccc|ccc|}
		\hline
		& \multicolumn{3}{c|}{Average Position Error (m)} & \multicolumn{3}{c|}{Average Velocity Error (m/s)} \\ 
		& InfoSSM & H-GPSSM & PRSSM & InfoSSM & H-GPSSM & PRSSM \\ \hline
		East-West & \bf 20.2375   & 24.6978   & 23.7322    & \bf 6.1574   & 11.2264    & 9.6225 \\ 
		North-South & 22.7680   & \bf 19.6529   & 25.6338    & \bf 6.4174   & 10.3157   & 10.6262 \\
		Horizontal & \bf 33.9305   & 34.7536   & 39.1680    & \bf 9.9167   & 16.9407   & 15.8674  \\
		\hline  
	\end{tabular}
\end{table}

\subsection{Application in Air Traffic Control Tracking}
After training InfoSSM, learned dynamics can be used for ATC tracking with filtering algorithms, as introduced in section \ref{sec:Filter}.
Since our InfoSSM only learned horizontal motion of maneuvering airplane, we additionally trained the GP dynamics for the vertical motion.
For the model simplicity, we train the vertical motion by the state-space model with single GP (i.e. PRSSM approach).

\begin{figure}[hbt!]
	\centering
	\subfigure[]{
		\includegraphics[width=0.55\columnwidth]{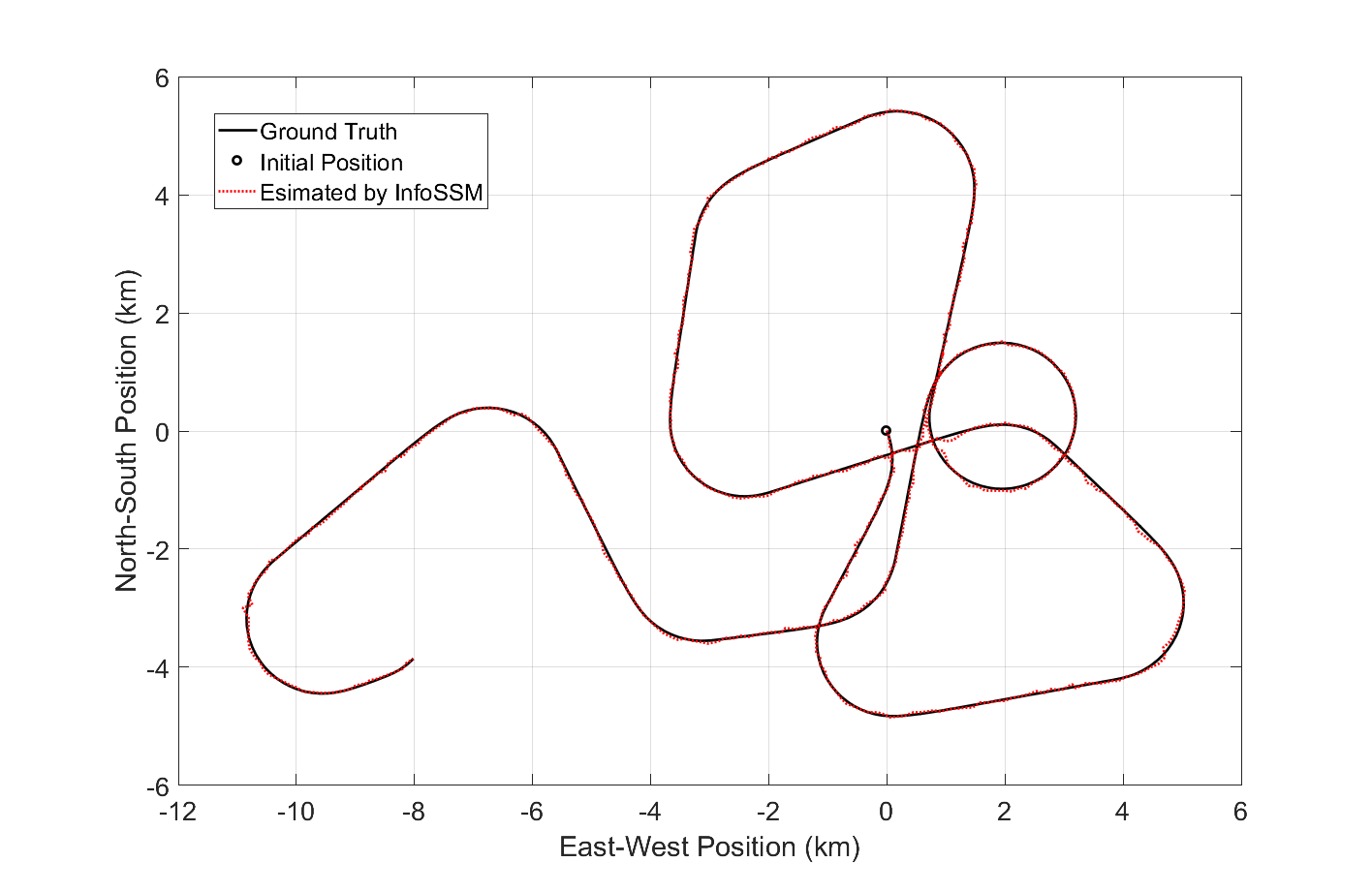}}
	\subfigure[]{
		\includegraphics[width=0.45\columnwidth]{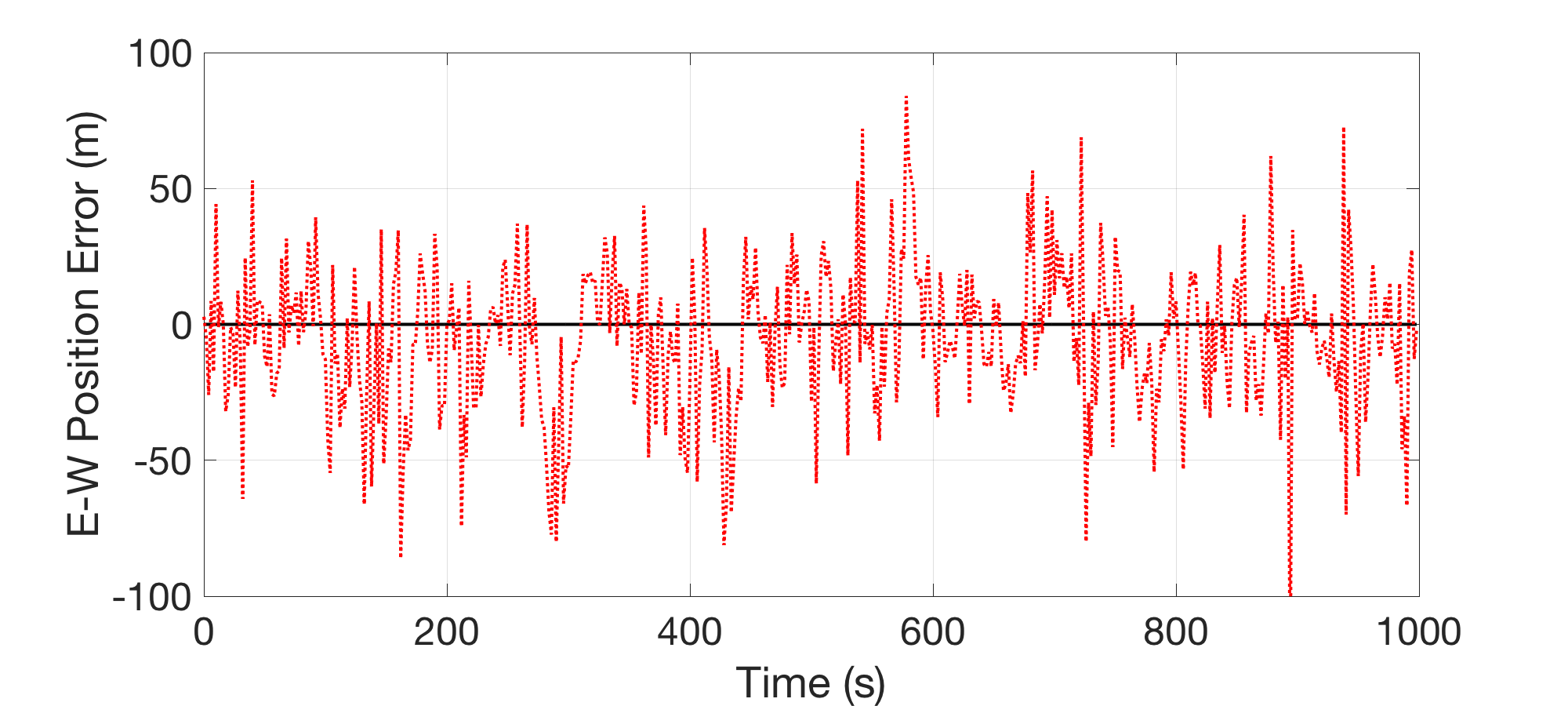}}
	\subfigure[]{
		\includegraphics[width=0.45\columnwidth]{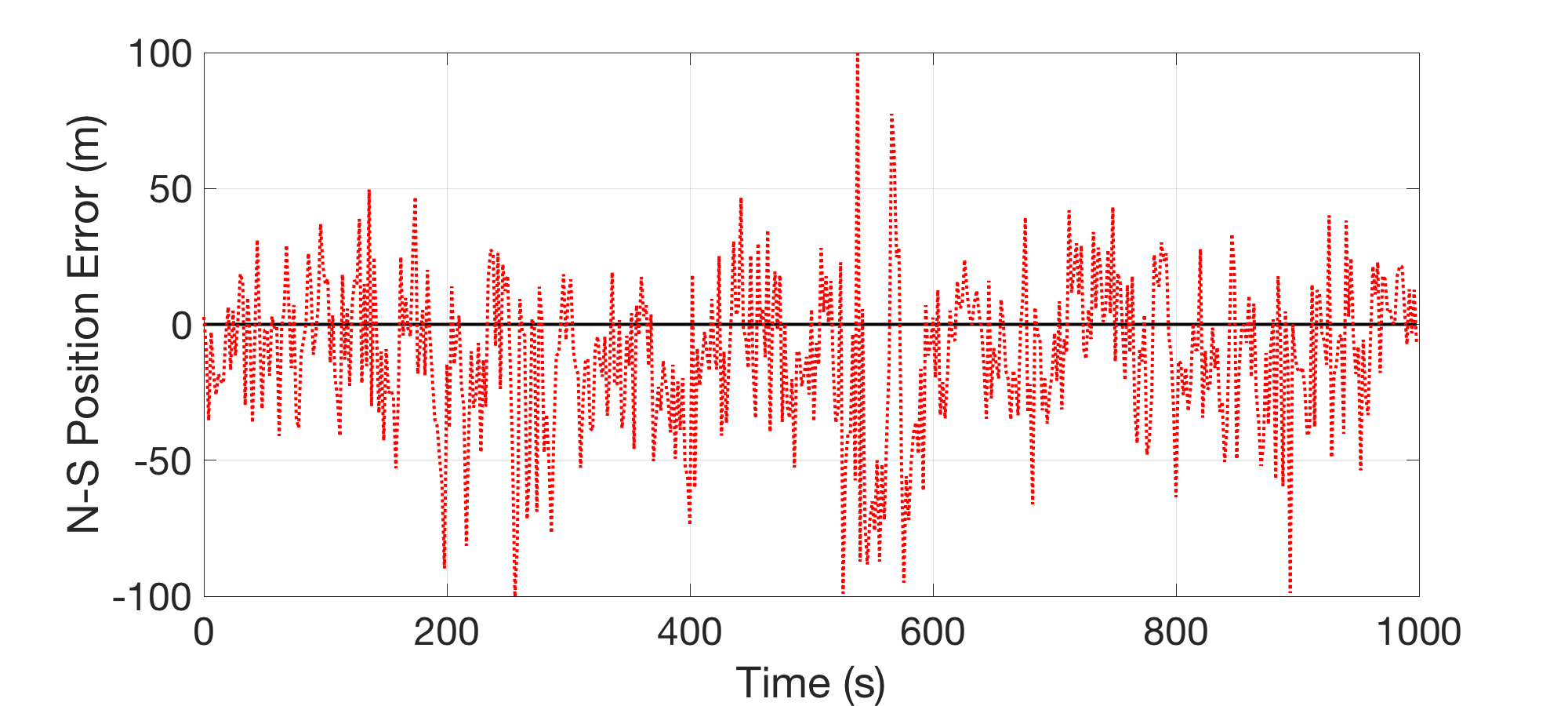}}
	\subfigure[]{
		\includegraphics[width=0.45\columnwidth]{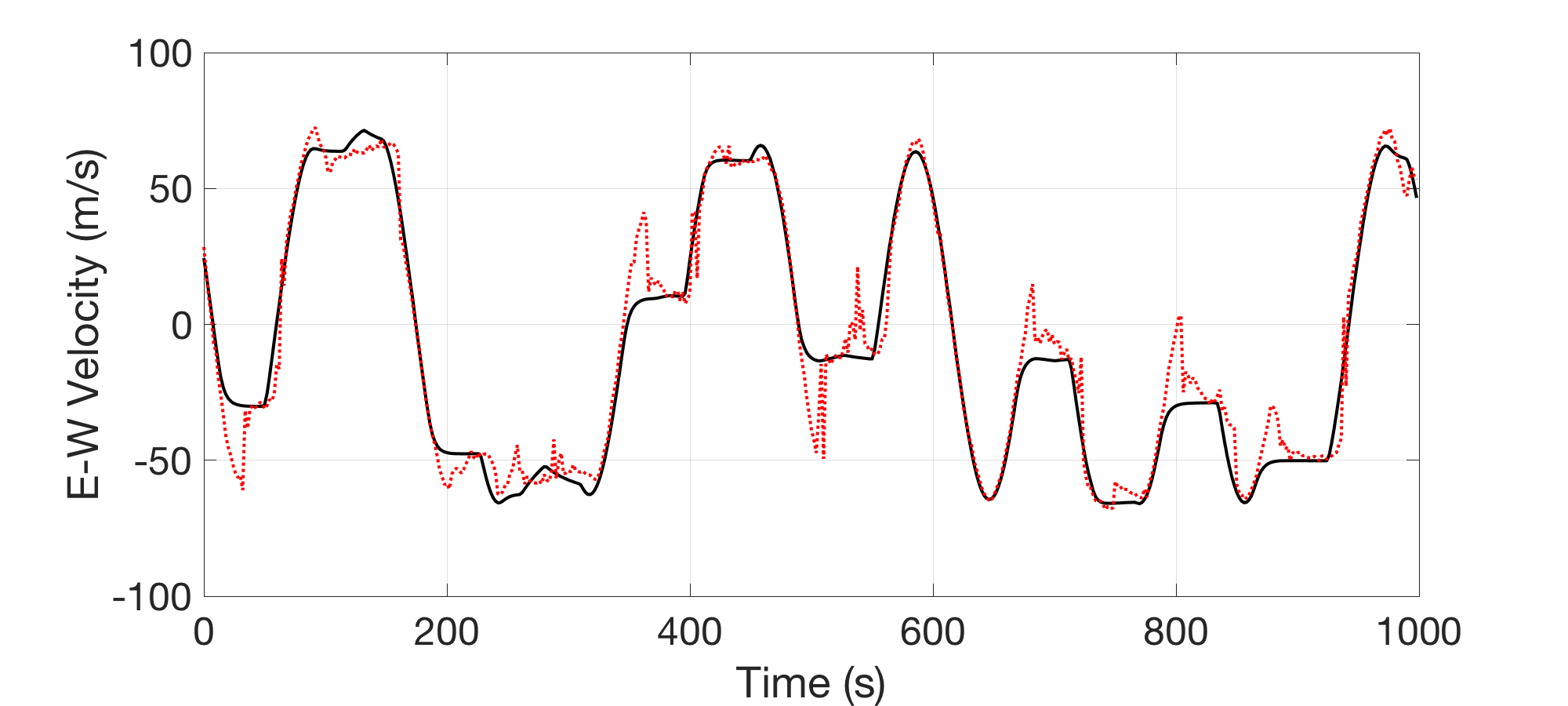}}
	\subfigure[]{
		\includegraphics[width=0.45\columnwidth]{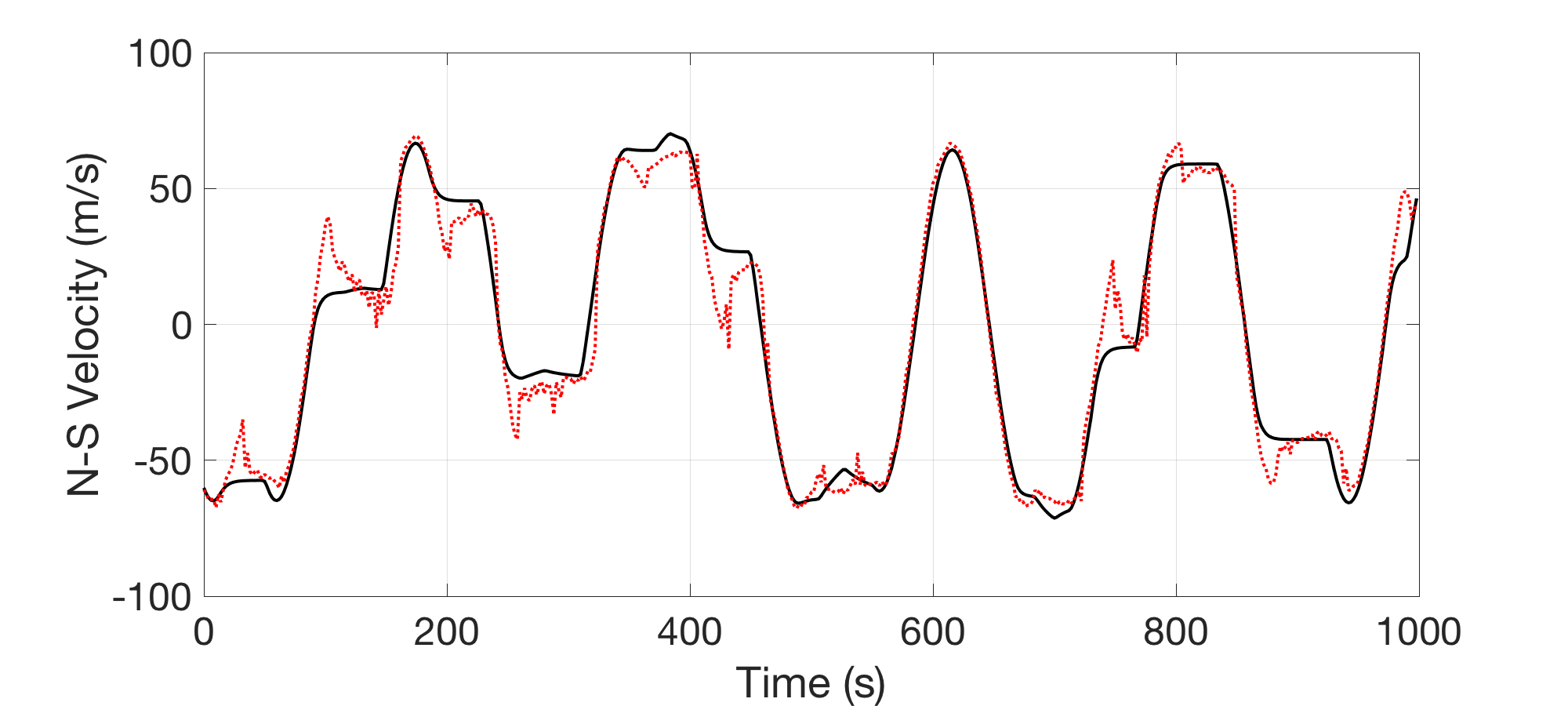}}
	\subfigure[]{
		\includegraphics[width=0.45\columnwidth]{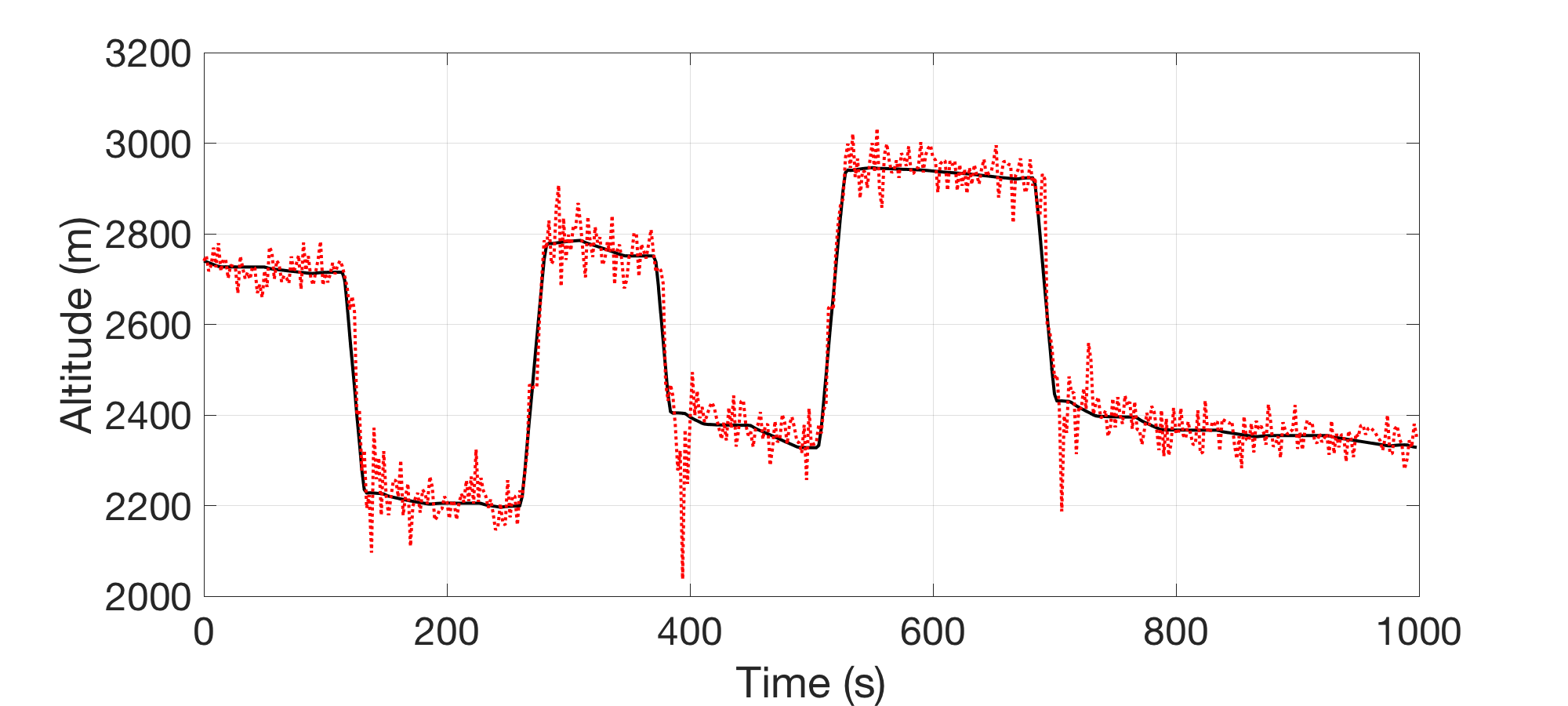}}
	\subfigure[]{
		\includegraphics[width=0.45\columnwidth]{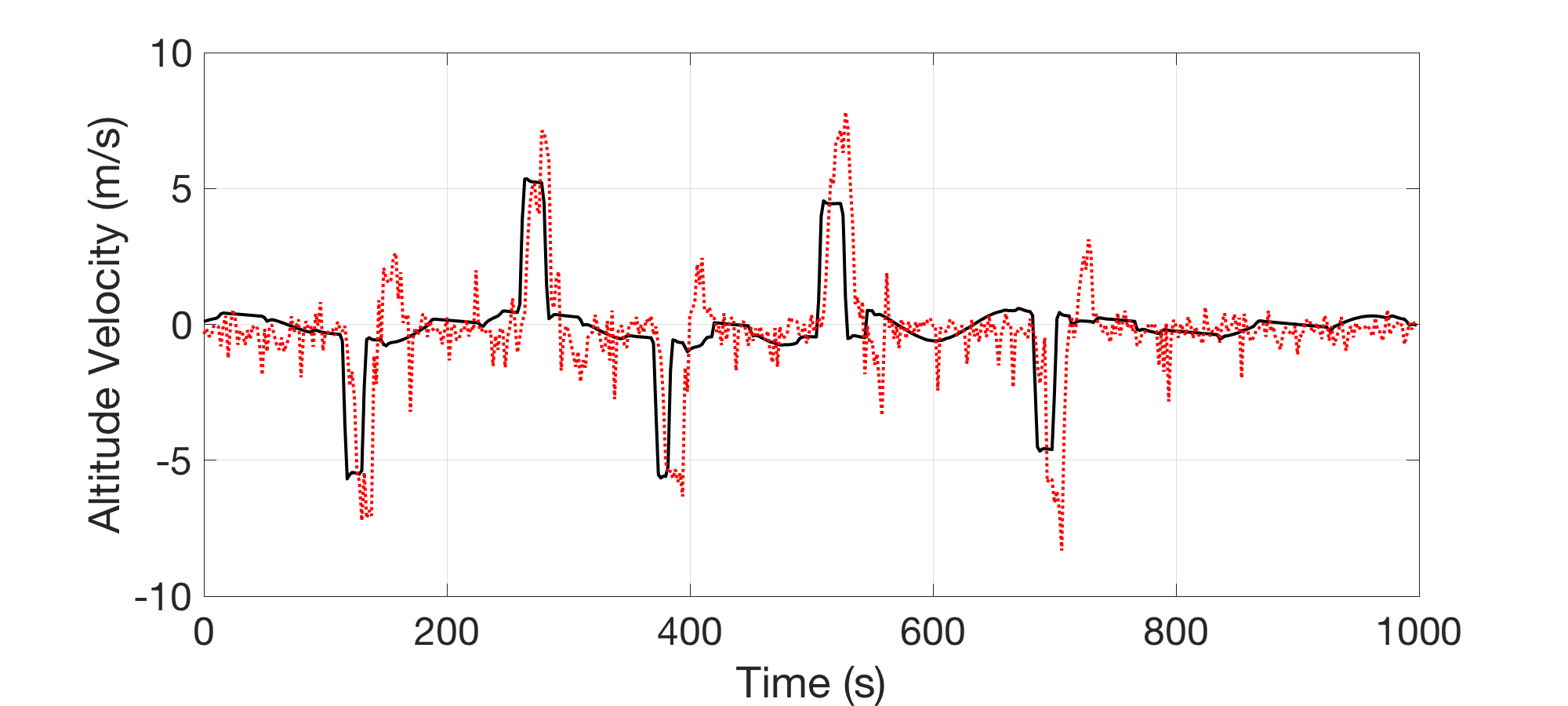}}
	\caption{ATC tracking results by using InfoSSM and particle filter. Ground truth and estimated states are colored by black and red, respectively. Average altitude position and velocity error are 33.03 m and 0.9830 m/s.}
	\label{fig:ATC}
\end{figure}

ATC simulation using X-Plane 11 was performed for 1000 seconds.
During the simulation, the 95\% accuracy bound (i.e. measurement error for the horizontal position is under 30 meters which corresponds to the NAC$_p$=9, and vertical position is under 45 meters \cite{mohleji2010modeling}) is considered for the measurement model.
Tracking results is provided in Table \ref{tab:ATC} and detail estimation results for InfoSSM is illustrated in Fig. \ref{fig:ATC}.
Although H-GPSSM shows better results for north-south position estimation, the overall horizontal estimation performance of InfoSSM is more accurate than other models.
Especially, InfoSSM shows much better accuracy in velocity estimation during turning motions which supports that the dynamics of InfoSSM is the most accurate among three models.
Further estimation results for H-GPSSM and PRSSM are in Fig. \ref{fig:ATC2} in Appendix D.

\newpage

\section{Conclusion}
In this paper, we presented the InfoSSM, an information theoretic extension of GPSSM that can effectively learn the multi-modal dynamical system.
To describe the multi-modal dynamics, we modeled multiple dynamics by using multiple GPs assigned by the latent dynamics code.
The inference of the latent state and dynamics code is performed via structured neural networks.
Unlike previous inference methods, InfoSSM could learn interpretable representation for dynamics by using the mutual information regularization without any supervision from a human.
The proposed model was evaluated with Dubins vehicle and high-fidelity flight simulator data and showed that the InfoSSM not only effectively represent multi-modal latent dynamics, but can reconstruct and predict the system.
Combining with Bayesian filtering algorithm, InfoSSM could be applied to ATC target tracking framework and gave decent performance.

\bibliographystyle{IEEEtran}
\bibliography{manuscript}

\newpage
\section*{Appendix A. Derivation of the MCO}
The MCO for deep Gaussian processes are previously introduced in \cite{park2018deep}. The model likelihood $\mathcal{P} \equiv p(\mathbf{Y}_{1:T}$) can be estimated by Monte Carlo estimator \cite{mnih2016variational}:
\begin{equation}
\mathcal{P} = \mathbb{E}_{q(\mathbf{X}_{1:T}^{(k)}, \mathbf{F}_{1:T-1}^{(k)}, \mathbf{U}, \mathbf{c}_{1:T-1}^{(k)})} \left[ \Big\langle \frac{p(\mathbf{Y}_{1:T},\mathbf{X}_{1:T}^{(k)}, \mathbf{F}_{1:T-1}^{(k)}, \mathbf{U}, \mathbf{c}_{1:T-1}^{(k)})}{q(\mathbf{X}_{1:T}^{(k)}, \mathbf{F}_{1:T-1}^{(k)}, \mathbf{U}, \mathbf{c}_{1:T-1}^{(k)})} \Big\rangle \right]
\end{equation}
where $\langle \cdot \rangle$ represents $\frac{1}{K}\sum_{k=1}^K(\cdot)$.
By using \eqref{eq:InfoSSM} and \eqref{eq:var},
\begin{align}
&\mathcal{P} = \mathbb{E}_{q(\mathbf{X}_{1:T}^{(k)}, \mathbf{F}_{1:T-1}^{(k)}, \mathbf{U}, \mathbf{c}_{1:T-1})} \left[ \Big\langle \frac{p(\mathbf{Y}_{1:T},\mathbf{X}_{1:T}^{(k)}, \mathbf{F}_{1:T-1}^{(k)}, \mathbf{U}, \mathbf{c}_{1:T-1}^{(k)})}{q(\mathbf{X}_{1:T}^{(k)}, \mathbf{F}_{1:T-1}^{(k)}, \mathbf{U}, \mathbf{c}_{1:T-1}^{(k)})} \Big\rangle \right] \nonumber \\
&= \mathbb{E}_{q(\mathbf{X}_{1:T}^{(k)}, \mathbf{F}_{1:T-1}^{(k)}, \mathbf{U}, \mathbf{c}_{1:T-1}^{(k)})} \left[ \Big\langle \prod_{t=1}^{T}p(\mathbf{Y}_t \mid \mathbf{X}_t^{(k)}) \cdot \frac{p(\mathbf{X}_1^{(k)})}{q(\mathbf{X}_1^{(k)})} \cdot \frac{p(\mathbf{U})}{q(\mathbf{U})} \cdot \frac{p(\mathbf{c}_1^{(k)})}{q(\mathbf{c}_1^{(k)})} \Big\rangle \right]  \nonumber \\
&= \mathbb{E}_{q(\mathbf{X}_{1:T}^{(k)}, \mathbf{F}_{1:T-1}^{(k)}, \mathbf{U}, \mathbf{c}_{1:T-1}^{(k)})} \left[ \frac{p(\mathbf{U})}{q(\mathbf{U})} \cdot \Big\langle \prod_{t=1}^{T}p(\mathbf{Y}_t \mid \mathbf{X}_t^{(k)}) \cdot \frac{p(\mathbf{X}_1^{(k)})}{q(\mathbf{X}_1^{(k)})} \cdot  \frac{p(\mathbf{c}_1^{(k)})}{q(\mathbf{c}_1^{(k)})} \Big\rangle \right]  \nonumber \\
&= \mathbb{E}_{q(\mathbf{U})} \left[ \frac{p(\mathbf{U})}{q(\mathbf{U})} \cdot
\mathbb{E}_{q(\mathbf{X}_{1:T}^{(k)}, \mathbf{F}_{1:T-1}^{(k)}, \mathbf{c}_{1:T-1}^{(k)})} \left[\Big\langle \prod_{t=1}^{T}p(\mathbf{Y}_t \mid \mathbf{X}_t^{(k)}) \cdot \frac{p(\mathbf{X}_1^{(k)})}{q(\mathbf{X}_1^{(k)})} \cdot  \frac{p(\mathbf{c}_1^{(k)})}{q(\mathbf{c}_1^{(k)})} \Big\rangle \right] \right]
\end{align}
By using Jensen's Inequality, the lower bound of marginal log likelihood is given by
\begin{align}
& \log \mathcal{P} \nonumber \\
& \ge \mathbb{E}_{q(\mathbf{U})} \left[ \log \frac{p(\mathbf{U})}{q(\mathbf{U})} \cdot
\mathbb{E}_{q(\mathbf{X}_{1:T}^{(k)}, \mathbf{F}_{1:T-1}^{(k)}, \mathbf{c}_{1:T-1}^{(k)})} \left[\Big\langle \prod_{t=1}^{T}p(\mathbf{Y}_t \mid \mathbf{X}_t^{(k)}) \cdot \frac{p(\mathbf{X}_1^{(k)})}{q(\mathbf{X}_1^{(k)})} \cdot  \frac{p(\mathbf{c}_1^{(k)})}{q(\mathbf{c}_1^{(k)})} \Big\rangle \right] \right] \nonumber \\
& = \mathbb{E}_{q(\mathbf{U})} \left[ \log \frac{p(\mathbf{U})}{q(\mathbf{U})} \right]
+ \mathbb{E}_{q(\mathbf{U})} \left[ \log \mathbb{E}_{q(\mathbf{X}_{1:T}^{(k)}, \mathbf{F}_{1:T-1}^{(k)}, \mathbf{c}_{1:T-1}^{(k)})} \left[ \Big\langle \prod_{t=1}^{T}p(\mathbf{Y}_t \mid \mathbf{X}_t^{(k)}) \cdot \frac{p(\mathbf{X}_1^{(k)})}{q(\mathbf{X}_1^{(k)})} \cdot  \frac{p(\mathbf{c}_1^{(k)})}{q(\mathbf{c}_1^{(k)})} \Big\rangle \right] \right] \nonumber \\
& \ge \mathbb{E}_{q(\mathbf{U})} \left[ \log \frac{p(\mathbf{U})}{q(\mathbf{U})} \right]
+ \mathbb{E}_{q(\mathbf{X}_{1:T}^{(k)}, \mathbf{c}_{1:T-1}^{(k)})} \left[ \log \Big\langle \prod_{t=1}^{T}p(\mathbf{Y}_t \mid \mathbf{X}_t^{(k)}) \cdot \frac{p(\mathbf{X}_1^{(k)})}{q(\mathbf{X}_1^{(k)})} \cdot  \frac{p(\mathbf{c}_1^{(k)})}{q(\mathbf{c}_1^{(k)})} \Big\rangle \right] \nonumber \\
&= \sum_{n=1}^{N} \mathbb{E}_{q(\mathbf{x}_{1:T,n}^{(k)}, \mathbf{c}_{1:T-1,n}^{(k)})} \left[ \log \Big\langle \prod_{t=1}^{T}p(\mathbf{y}_t \mid \mathbf{x}_{t,n}^{(k)}) \cdot \frac{p(\mathbf{x}_{1,n}^{(k)})}{q(\mathbf{x}_{1,n}^{(k)})} \cdot \frac{p(c_{1,n}^{(k)})}{q(c_{1,n}^{(k)})} \Big\rangle \right]
- \sum_{l=1}^{L} KL(q(\mathbf{U}_l)||p(\mathbf{U}_l)) .
\end{align}
Hence the MCO defined in \eqref{eq:MCO} is the lower bound of the model marginal log-likelihood.

Remark that, the MCO becomes equivalent to the ELBO when $K=1$.
In fact, defined MCO is equivalent to the IWAE bound, thus it gets monotonically tighter as the number of sample size $K$ increases \cite{burda2015IWAE}.
Hence, the MCO achieves a tighter lower bound than the traditional ELBO:
\begin{equation}
\mathcal{L}(\mathbf{Y}_{1:T})=\mathcal{L}^1(\mathbf{Y}_{1:T}) \le \mathcal{L}^K(\mathbf{Y}_{1:T}) \le \log \mathcal{P}.
\end{equation}

\section*{Appendix B. Derivation of the Variational Lower Bound of Mutual Information}
\begin{align}
&\mathcal{I}(\mathbf{c}_{1:T-1}; \mathcal{G}(x_1, \mathbf{c}_{1:T-1})) 
= H(\mathbf{c}_{1:T-1}) - H(\mathbf{c}_{1:T-1} \mid \mathcal{G}(x_1, \mathbf{c}_{1:T-1})) \nonumber \\
&= H(\mathbf{c}_{1:T-1}) + \mathbb{E}_{\mathbf{y}_{1:T} \sim \mathcal{G}(x_1, \mathbf{c}_{1:T-1})} \Big[ \mathbb{E}_{\mathbf{c}'_{1:T-1} \sim p(\mathbf{c}_{1:T-1} \mid \mathbf{y}_{1:T})} [\log p(\mathbf{c}'_{1:T-1} ; \mathbf{y}_{1:T})] \Big] \nonumber \\
&= H(\mathbf{c}_{1:T-1}) + \mathbb{E}_{\mathbf{y}_{1:T} \sim \mathcal{G}(x_1, \mathbf{c}_{1:T-1})} \Big[ KL(p(\cdot \mid \mathbf{y}_{1:T}) ~ || ~ q(\cdot \mid \mathbf{y}_{1:T})) + \mathbb{E}_{\mathbf{c}'_{1:T-1} \sim p(\mathbf{c}_{1:T-1} \mid \mathbf{y}_{1:T})} [\log q(\mathbf{c}'_{1:T-1} ; \mathbf{y}_{1:T})] \Big] \nonumber \\
&\ge H(\mathbf{c}_{1:T-1}) + \mathbb{E}_{\mathbf{y}_{1:T} \sim \mathcal{G}(x_1, \mathbf{c}_{1:T-1})} \Big[ \mathbb{E}_{\mathbf{c}'_{1:T-1} \sim p(\mathbf{c}_{1:T-1} \mid \mathbf{y}_{1:T})} [\log p(\mathbf{c}'_{1:T-1} ; \mathbf{y}_{1:T})] \Big] \nonumber \\
&= H(\mathbf{c}_{1:T-1}) + \mathbb{E}_{\mathbf{c}_{1:T-1} \sim p(\mathbf{c}_{1:T-1}), ~ \mathbf{y}_{1:T} \sim \mathcal{G}(x_1, \mathbf{c}_{1:T-1})} \Big[ \log q(\mathbf{c}_{1:T-1} ; \mathbf{y}_{1:T}) \Big] \nonumber \\
&= \mathbb{E}_{\mathbf{c}_{1:T-1} \sim p(\mathbf{c}_{1:T-1})}\Big[-\log p(\mathbf{c}_{1:T-1}) \Big] + \mathbb{E}_{\mathbf{c}_{1:T-1} \sim p(\mathbf{c}_{1:T-1}), ~ \mathbf{y}_{1:T} \sim \mathcal{G}(x_1, \mathbf{c}_{1:T-1})} \Big[ \log q(\mathbf{c}_{1:T-1} ; \mathbf{y}_{1:T}) \Big] \nonumber \\
&= \mathbb{E}_{\mathbf{c}_{1:T-1} \sim p(\mathbf{c}_{1:T-1}), ~ \mathbf{y}_{1:T} \sim \mathcal{G}(x_1, \mathbf{c}_{1:T-1})} \Big[ \log q(\mathbf{c}_{1:T-1} ; \mathbf{y}_{1:T}) - \log p(\mathbf{c}_{1:T-1}) \Big] \nonumber \\
&\approx \mathbb{E}_{\mathbf{c}_{1:T-1} \sim p(\mathbf{c}_{1:T-1}), ~ \mathbf{y}_{1:T} \sim \mathcal{G}(x_1, \mathbf{c}_{1:T-1})} \Big[ \log \frac{q(\mathbf{c}_{1} ; \mathbf{y}_{1:T})\prod_{t=1}^{T-2} p(c_{t+1} \mid c_{t})}{p(\mathbf{c}_{1})\prod_{t=1}^{T-2} p(c_{t+1} \mid c_{t})} \Big] \nonumber \\
&= \mathbb{E}_{\mathbf{c}_{1:T-1} \sim p(\mathbf{c}_{1:T-1}), ~ \mathbf{y}_{1:T} \sim \mathcal{G}(x_1, \mathbf{c}_{1:T-1})} \Big[ \log q(\mathbf{c}_{1} ; \mathbf{y}_{1:T}) - \log p(\mathbf{c}_{1}) \Big] \nonumber \\
&= \mathbb{E}_{\mathbf{c}_{1} \sim p(\mathbf{c}_{1})}\Big[-\log p(\mathbf{c}_{1}) \Big] + \mathbb{E}_{\mathbf{c}_{1:T-1} \sim p(\mathbf{c}_{1:T-1}), ~ \mathbf{y}_{1:T} \sim \mathcal{G}(x_1, \mathbf{c}_{1:T-1})} \Big[ \log q(\mathbf{c}_{1} ; \mathbf{y}_{1:T}) \Big] \nonumber \\
&= H(c_1) + \mathbb{E}_{\mathbf{c}_{1:T-1} \sim p(\mathbf{c}_{1:T-1}), ~ \mathbf{y}_{1:T} \sim \mathcal{G}(x_1, \mathbf{c}_{1:T-1})} \Big[ \log q(c_1 ; \mathbf{y}_{1:T}) \Big].
\end{align}

\section*{Appendix C. Experiment Details}
\subsection*{C.1. Shift and Rotation Invariant Inference Structures}
In the present work, we are particularly focusing on the state-space models in the Cartesian coordinate.
Dynamics in Cartesian coordinate has two general characteristics; physics are \textit{invariant} to shift and rotation in the horizontal space.
As such, we hope networks output the same dynamics code even if we input the shifted and rotated trajectories.
To embed such characteristics, we propose a shift and rotation invariant inference structure via simple idea;
we shift the initial position of the input trajectory to the origin and rotate the shifted trajectory with random angle:
\begin{align}
\mathbf{y}_{1:T,n} &\leftarrow \mathbf{\Phi} (\mathbf{y}_{1:T,n} - \mathbf{y}_{1,n}) \nonumber \\
q(c_{1,n}) &= Cat\Big(c_{1,n} ; Q_\phi (\mathbf{y}_{1:T, n})\Big)
\end{align}
where $\mathbf{\Phi}$ is function rotates the input trajectory with random angle in horizontal space.
After the inference 
This also can be interpreted as data augmentation techniques.

\begin{figure}[hbt!]
	\centering
	\includegraphics[width=0.85\textwidth]{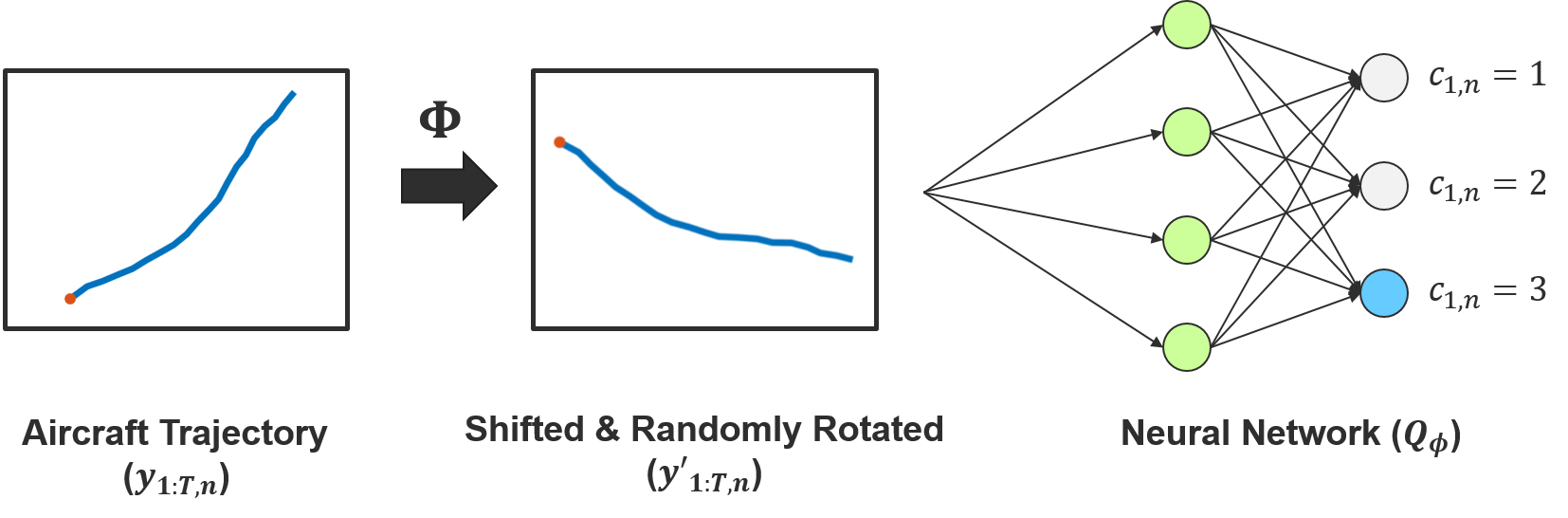}
	\caption{The inference structure for the initial code of InfoSSM}
\end{figure}

\subsection*{C.2. Dubins Airplane}
For both Dubins example, \textit{canonical} state-space model with $\mathbf{x} = [\mathbf{p}_x, \mathbf{p}_y, \mathbf{v}_x, \mathbf{v}_y]$ is constructed as
\begin{align}
[\dot{\mathbf{p}}_x, \dot{\mathbf{p}}_y] &= [\mathbf{v}_x, \mathbf{v}_y] \nonumber \\
[\dot{\mathbf{v}}_x, \dot{\mathbf{v}}_y] &= \sum_{l=1}^L p(\mathbf{c}=l) [\mathcal{GP}_l(\mathbf{v}_x, \mathbf{v}_y) + \boldsymbol{\epsilon}_{f,l} ]\nonumber \\
\mathbf{y} &= [\mathbf{p}_x, \mathbf{p}_y] + \boldsymbol{\epsilon}_g
\end{align}
where $\boldsymbol{\epsilon}_{f,l} \sim \mathcal{N}(\mathbf{0}, \mbox{diag}[\sigma_{v_x,l}^2, \sigma_{v_y,l}^2])$ and $\boldsymbol{\epsilon}_g \sim \mathcal{N}(\mathbf{0}, \mbox{diag}[\sigma_{p_x}^2, \sigma_{p_y}^2])$ that are both trainable variables.
GPSSMs are constructed with $M=20$, $K=4$, $T=20$, and $\delta t=0.1$.
For InfoSSM and H-GPSSM, three GPs are used ($L=3$).
We set $\lambda = NT$ for InfoSSM.

\subsection*{C.3. High-Fidelity Flight Simulator}
For both high-fidelity flight simulator example, \textit{canonical} state-space model with $\mathbf{x} = [\mathbf{p}_x, \mathbf{p}_y, \mathbf{v}_x, \mathbf{v}_y, \mathbf{a}_x, \mathbf{a}_y]$ is constructed as
\begin{align}
[\dot{\mathbf{p}}_x, \dot{\mathbf{p}}_y] &= [\mathbf{v}_x, \mathbf{v}_y] \nonumber \\
[\dot{\mathbf{v}}_x, \dot{\mathbf{v}}_y] &= [\mathbf{a}_x, \mathbf{a}_y] \nonumber \\
[\dot{\mathbf{a}}_x, \dot{\mathbf{a}}_y] &= \sum_{l=1}^L p(\mathbf{c}=l) [\mathcal{GP}_l(\mathbf{v}_x, \mathbf{v}_y, \mathbf{a}_x, \mathbf{a}_y) + \boldsymbol{\epsilon}_{f,l}] \nonumber \\
\mathbf{y} &= [\mathbf{p}_x, \mathbf{p}_y] + \boldsymbol{\epsilon}_g
\end{align}
where $\boldsymbol{\epsilon}_{f,l} \sim \mathcal{N}(\mathbf{0}, \mbox{diag}[\sigma_{a_x,l}^2, \sigma_{a_y,l}^2])$ are trainable variables, and $\boldsymbol{\epsilon}_g \sim \mathcal{N}(\mathbf{0}, \mbox{diag}[30^2, 30^2])$ in SI units.
GPSSMs are constructed with $M=50$, $K=8$, $T=20$, and $\delta t=2$.
For InfoSSM and H-GPSSM, three GPs are used ($L=3$).
We set $\lambda = 10NT$ for InfoSSM.

\begin{algorithm}[b!]
	\caption{Particle Filter Implementation}
	\label{al:PF}
	\begin{algorithmic}[1]
		\REQUIRE $\{\mathbf{x}^{(k)}_{t-1}, \mathbf{c}^{(k)}_{t-1}, w_{t-1}^{(k)}\}$: latent states, and $\mathbf{y}_{t}$: observation
		\FOR {$k = 1$ to $K$}
		\STATE Sample $\tilde{\mathbf{c}}^{(k)}_{t}$ from $\mathbf{c}^{(k)}_{t-1}$, using \eqref{eq:jump}.
		\STATE Sample $\tilde{\mathbf{x}}^{(k)}_{t}$ from $\mathbf{x}^{(k)}_{t-1}$, using \eqref{eq:propagate}.
		\STATE Compute the likelihood, $p(\tilde{\mathbf{x}}_t^{(k)} \mid \mathbf{y}_t)$, using the observation model.
		\STATE Update the weight, $w_t^{(k)}$, using \eqref{eq:weight}.
		\ENDFOR
		\STATE Compute the ESS: $\Big(\sum_{k=1}^K (w_t^{(k)})^2\Big)^{-1}$.
		\IF {$ESS \le K/2$}
		\FOR {$k = 1$ to $K$}
		\STATE Sample $i \sim Cat(\{w_t^{(1)}, \cdots, w_t^{(K)}\}) $.
		\STATE $\mathbf{x}^{(k)}_{t}, \mathbf{c}^{(k)}_{t} \leftarrow \tilde{\mathbf{x}}^{(i)}_{t}, \tilde{\mathbf{c}}^{(i)}_{t}$
		\ENDFOR
		\STATE $w_t^{(k)} \leftarrow 1/K$ for $\{k = 1, \cdots, K\}$.
		\ELSE
		\STATE $\mathbf{x}^{(k)}_{t}, \mathbf{c}^{(k)}_{t} \leftarrow \tilde{\mathbf{x}}^{(k)}_{t}, \tilde{\mathbf{c}}^{(k)}_{t}$ for $\{k = 1, \cdots, K\}$.
		\ENDIF
	\end{algorithmic}
\end{algorithm}

\subsection*{C.3. Air Traffic Control Tracking}
Horizontal motion model is built by InfoSSM explained in the above.
Similarly, state-space for vertical motion model is constructed as

\begin{align}
\dot{\mathbf{p}}_h &= \mathbf{v}_h \nonumber \\
\dot{\mathbf{v}}_h &= \mathbf{a}_h \nonumber \\
\dot{\mathbf{a}}_h &= \mathcal{GP}_h (\mathbf{v}_h, \mathbf{a}_h) + \boldsymbol{\epsilon}_{f,h} \nonumber \\
\mathbf{y}_h &= [\mathbf{p}_z] + \boldsymbol{\epsilon}_{g,h}
\end{align}
where $\boldsymbol{\epsilon}_{f,h} \sim \mathcal{N}(0, \sigma_{a_h}^2)$ are trainable variables, and $\boldsymbol{\epsilon}_{g,h} \sim \mathcal{N}(0, 45^2)$ in SI units.
GPSSMs are constructed with $M=50$, $K=8$, $T=20$, and $\delta t=2$.
For PF filtering, implemented as algorithm \ref{al:PF}, total 512 particles are used.
During target tracking, transition matrix is fixed as follows:
\begin{equation}
P_{i,j} = 
\begin{cases}
0.9,  & \text{if } i=j\\
0.05, & \text{otherwise}
\end{cases}
\end{equation}

\section*{D. Further Results}

\begin{figure}[h!]
	\centering
	\subfigure[InfoSSM ($\lambda > 0$)]{
		\includegraphics[width=1.0\columnwidth]{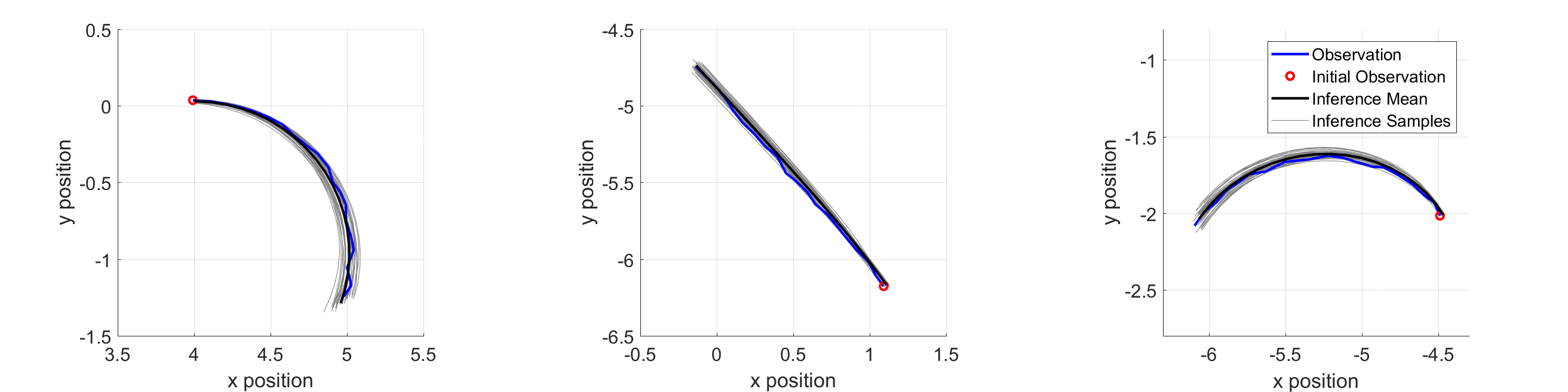}}
	\subfigure[H-GPSSM ($\lambda = 0$)]{
		\includegraphics[width=1.0\columnwidth]{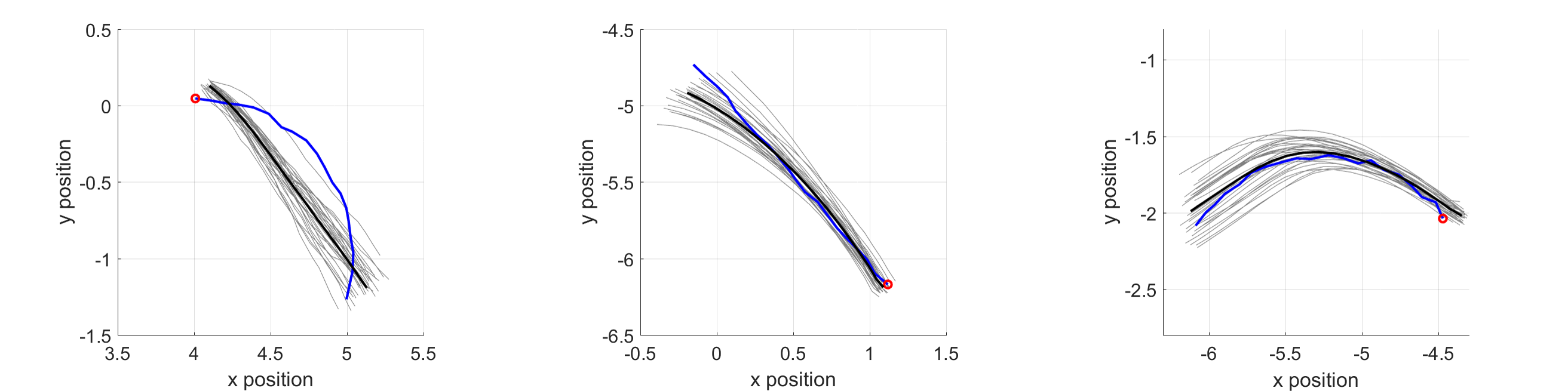}}
	\subfigure[PRSSM]{
		\includegraphics[width=1.0\columnwidth]{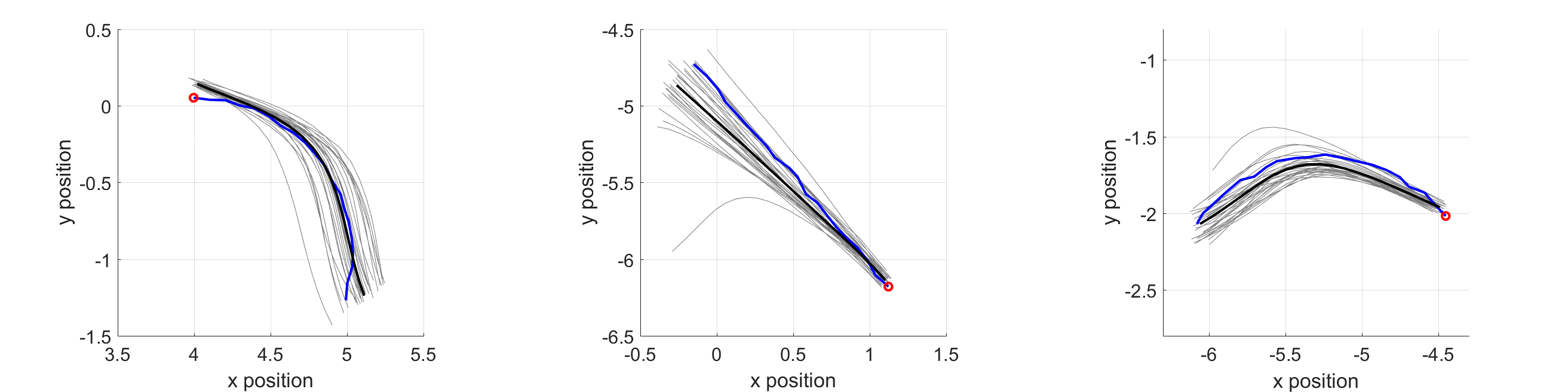}}
	\caption{Dubins: Reconstruction results at three different cases. From the left, let case 1, 2, and 3. The RMSE and log-likelihood of each result are shown in Table \ref{tab:reconD}.}
	\label{fig:reconD}
\end{figure}

\begin{table} [h!]
	\caption{Dubins: The RMSE and Log-likelihood of Test Cases in Fig. \ref{fig:reconD}}
	\centering
	\label{tab:reconD}
	\begin{tabular}{|l|ccc|ccc|}
		\hline
		& \multicolumn{3}{c|}{RMSE} & \multicolumn{3}{c|}{log-likelihood} \\ 
		& InfoSSM  & H-GPSSM & PRSSM & InfoSSM & H-GPSSM & PRSSM \\ \hline
		Case 1. Right & \bf 0.6767 & 2.8916 & 1.0623 & \bf 50.0152 & -32.6587 & 31.6831 \\
		Case 2. Straight & \bf 0.4091 & 1.6435 & 2.0696 & \bf 51.3238 & 30.6983 & 11.4390 \\
		Case 3. Left & \bf 0.7263 & 1.4932 & 1.5680 & \bf 35.8378 & -10.6636 & -27.3243 \\ \hline
	\end{tabular}
\end{table}

\begin{figure}[h!]
	\centering
	\subfigure[InfoSSM ($\lambda > 0$)]{
		\includegraphics[width=1.0\columnwidth]{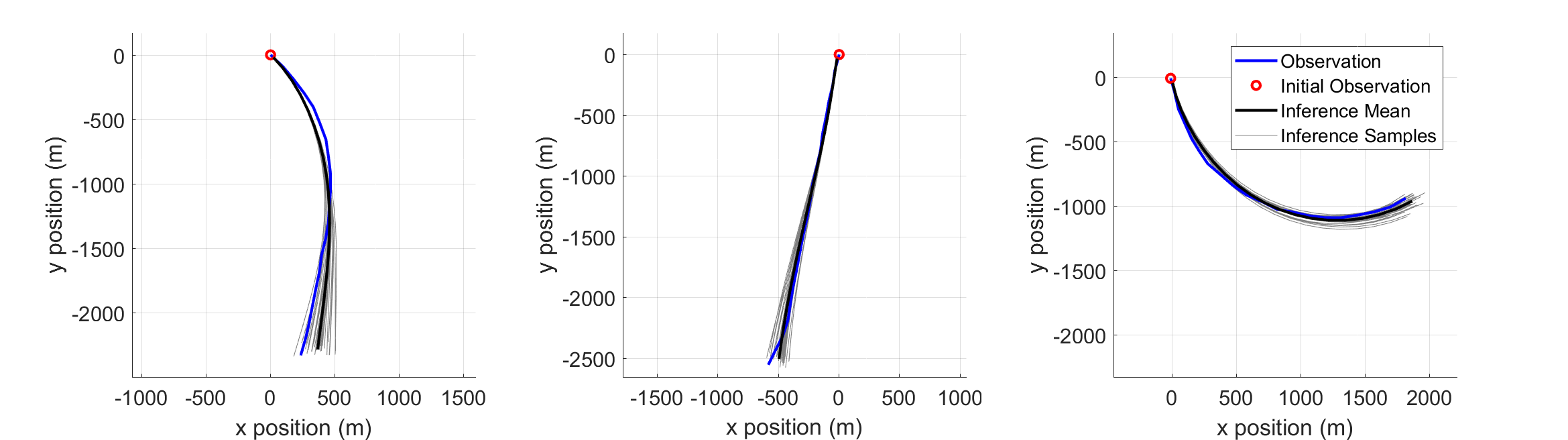}}
	\subfigure[H-GPSSM ($\lambda = 0$)]{
		\includegraphics[width=1.0\columnwidth]{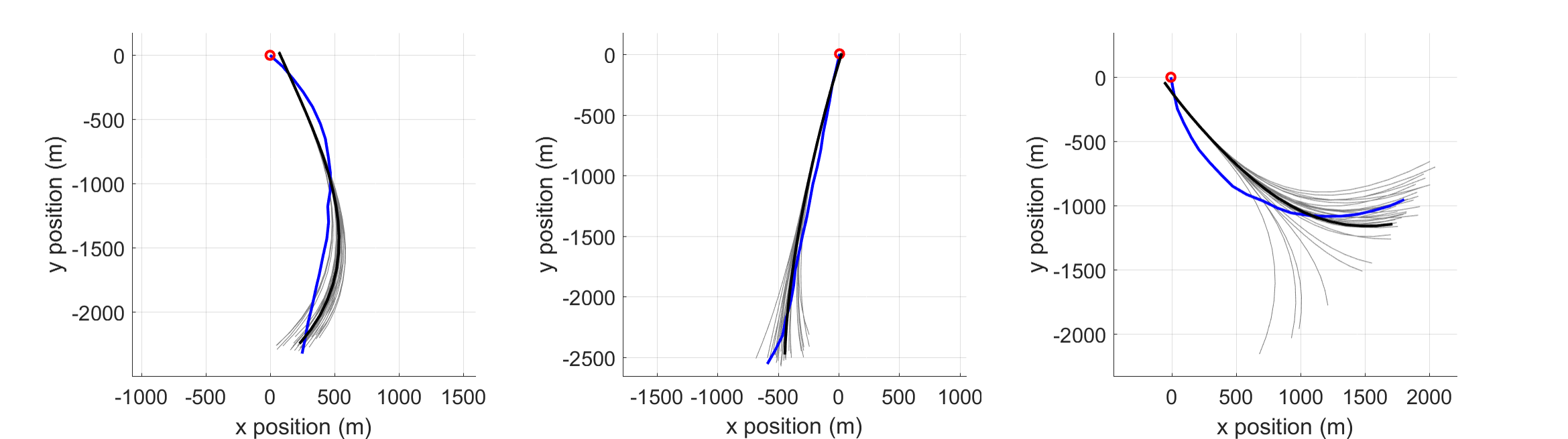}}
	\subfigure[PRSSM]{
		\includegraphics[width=1.0\columnwidth]{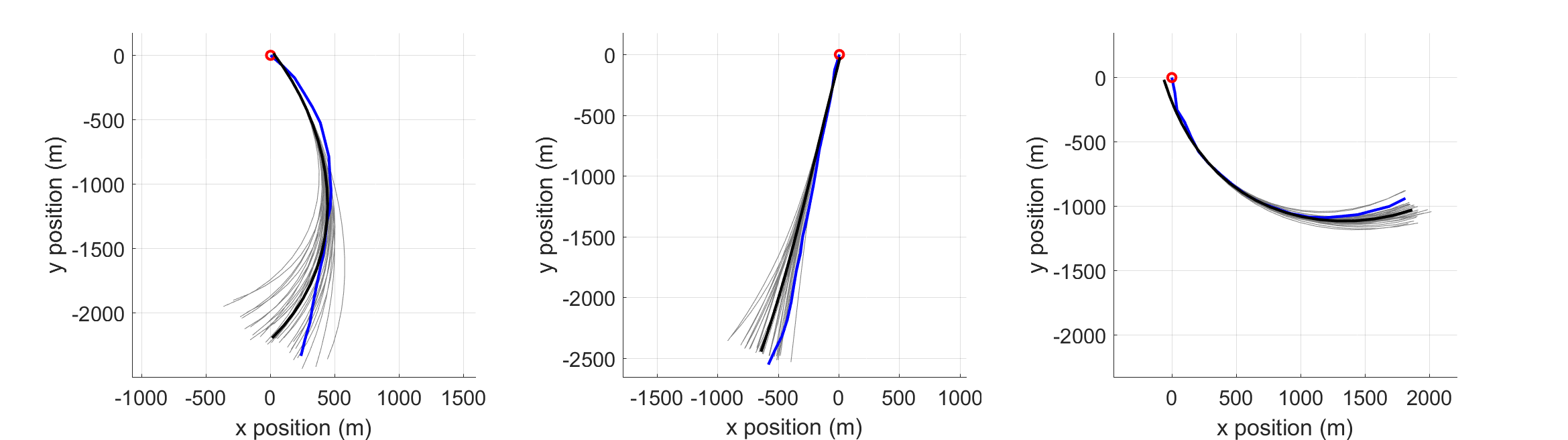}}
	\caption{Cessna: Reconstruction results at three different cases. From the left, let case 1, 2, and 3. The RMSE and log-likelihood of each result are shown in Table \ref{tab:reconC}.}
	\label{fig:reconC}
\end{figure}

\begin{table} [h!]
	\caption{Cessna: The RMSE and Log-likelihood of Test Cases in Fig. \ref{fig:reconC}}
	\centering
	\label{tab:reconC}
	\begin{tabular}{|l|ccc|ccc|}
		\hline
		& \multicolumn{3}{c|}{RMSE ($\times 10^3$)} & \multicolumn{3}{c|}{log-likelihood ($\times 10^3$)} \\ 
		& InfoSSM  & H-GPSSM & PRSSM & InfoSSM & H-GPSSM & PRSSM \\ \hline
		Case 1. Right & \bf 1.015 & 1.4592 & 1.2521 & \bf -0.4720 & -1.7942 & -0.4848 \\
		Case 2. Straight & \bf 0.7565 & 1.172 & 1.4298 & \bf -0.1409 & -0.2579 & -0.2390 \\
		Case 3. Left & \bf 0.7157 & 2.0995 & 0.8766 & \bf -0.1959 & -0.3853 & -1.1422 \\ \hline
	\end{tabular}
\end{table}

\begin{figure}[hbt!]
	\centering
	\subfigure[H-GPSSM]{
		\includegraphics[width=0.45\columnwidth]{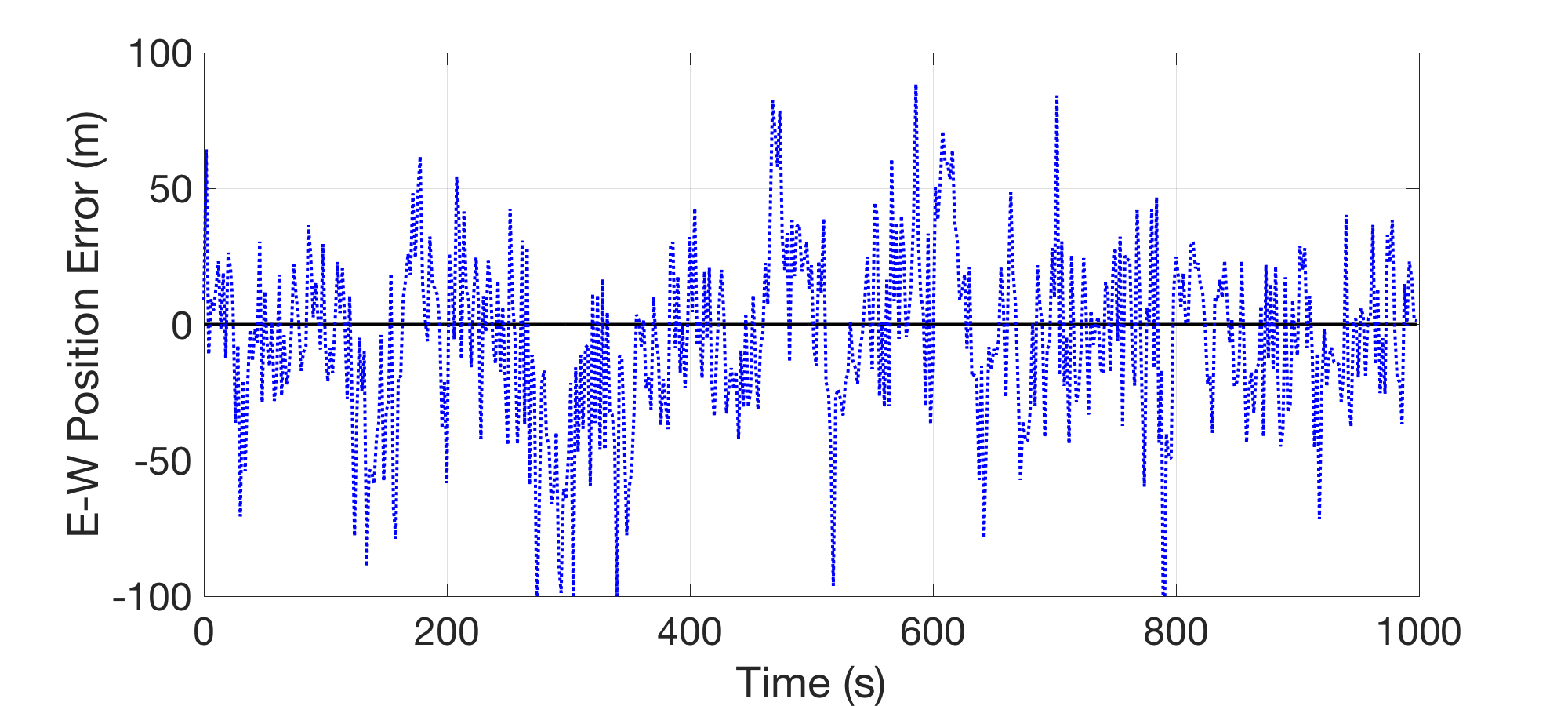}}
	\subfigure[PRSSM]{
		\includegraphics[width=0.45\columnwidth]{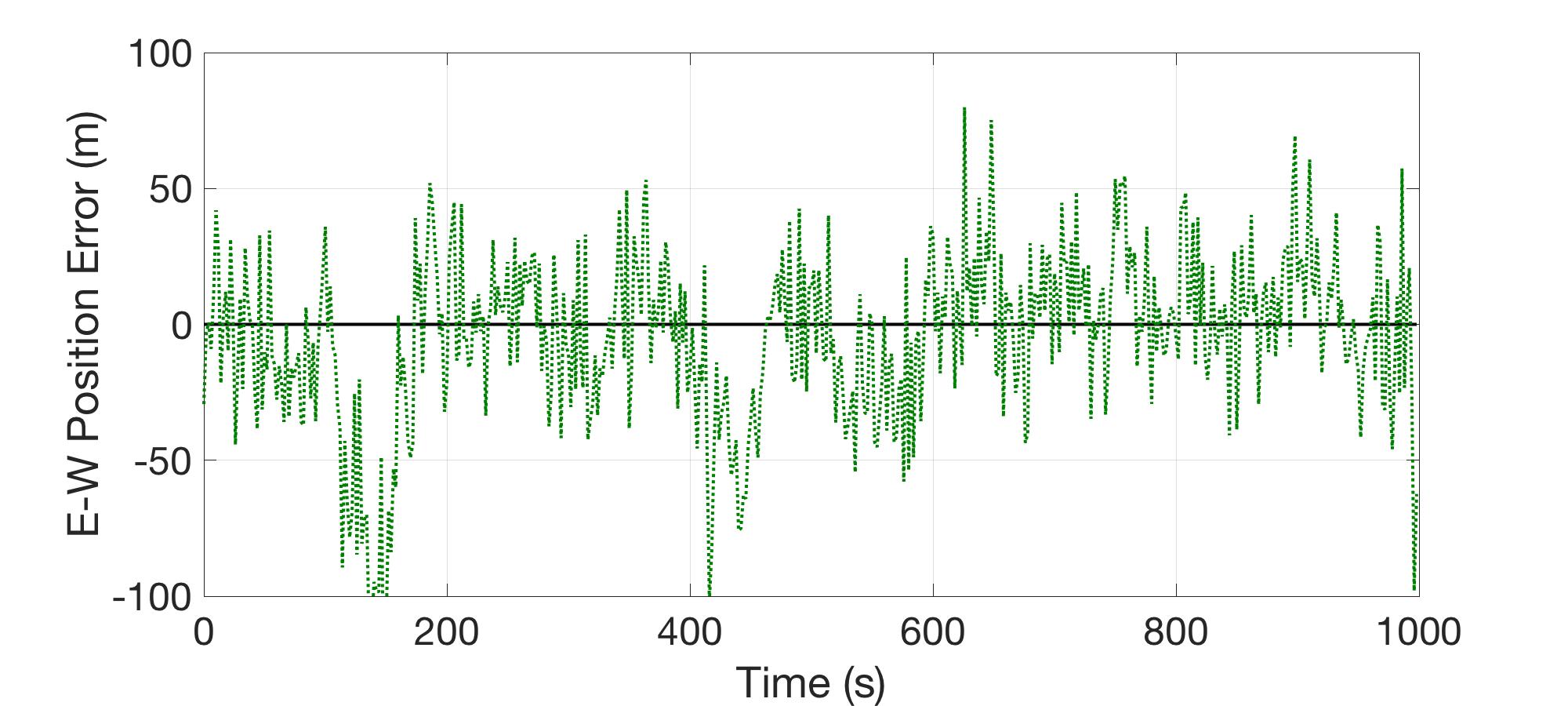}}
	\subfigure[H-GPSSM]{
		\includegraphics[width=0.45\columnwidth]{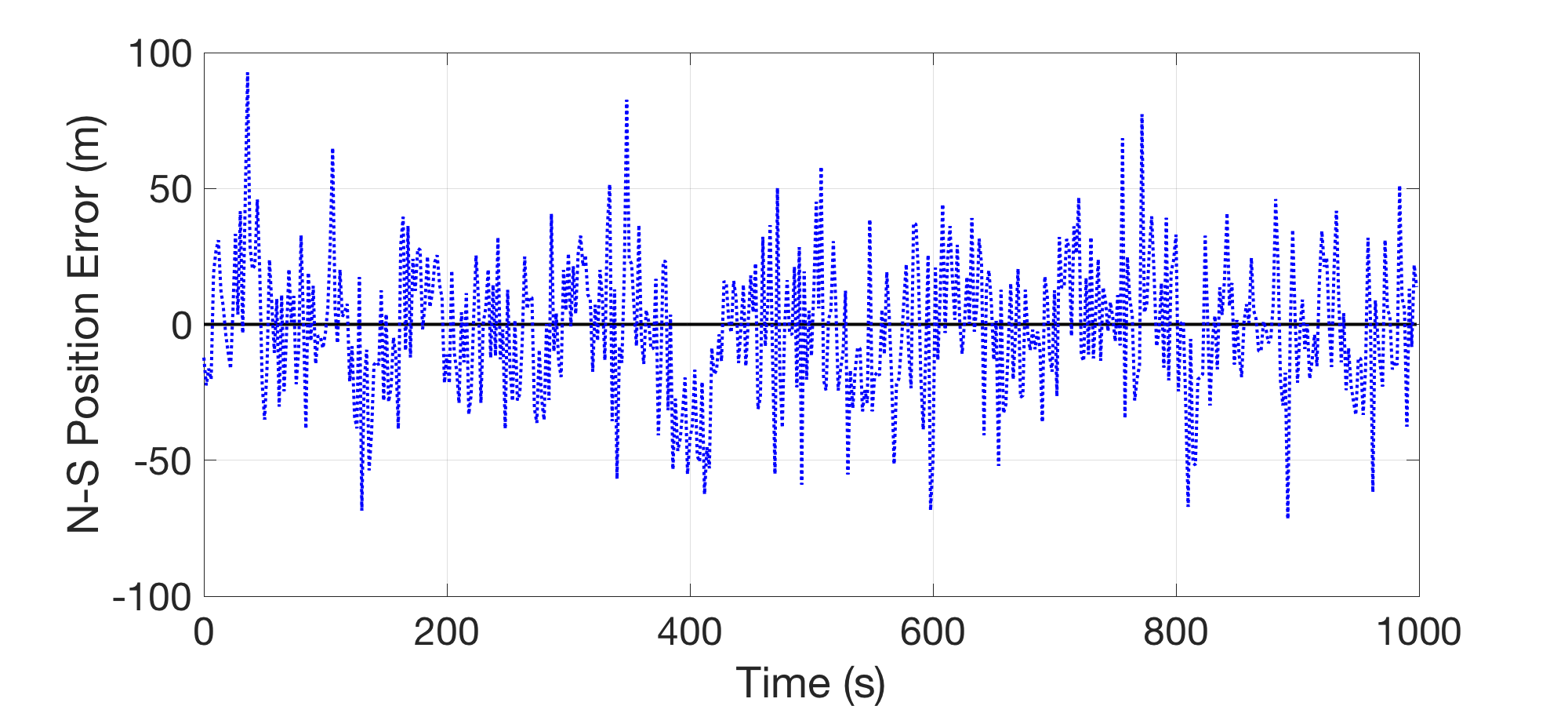}}
	\subfigure[PRSSM]{
		\includegraphics[width=0.45\columnwidth]{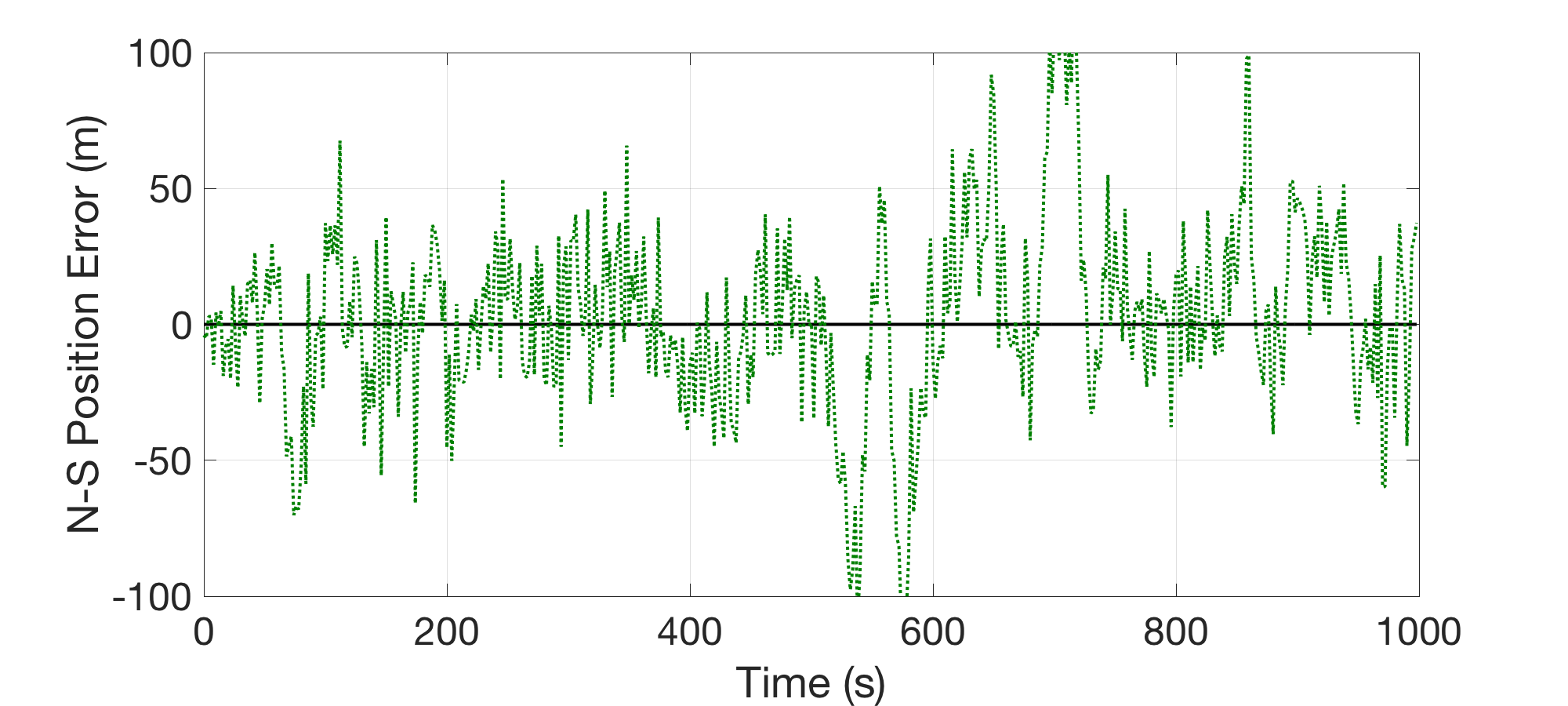}}
	\subfigure[H-GPSSM]{
		\includegraphics[width=0.45\columnwidth]{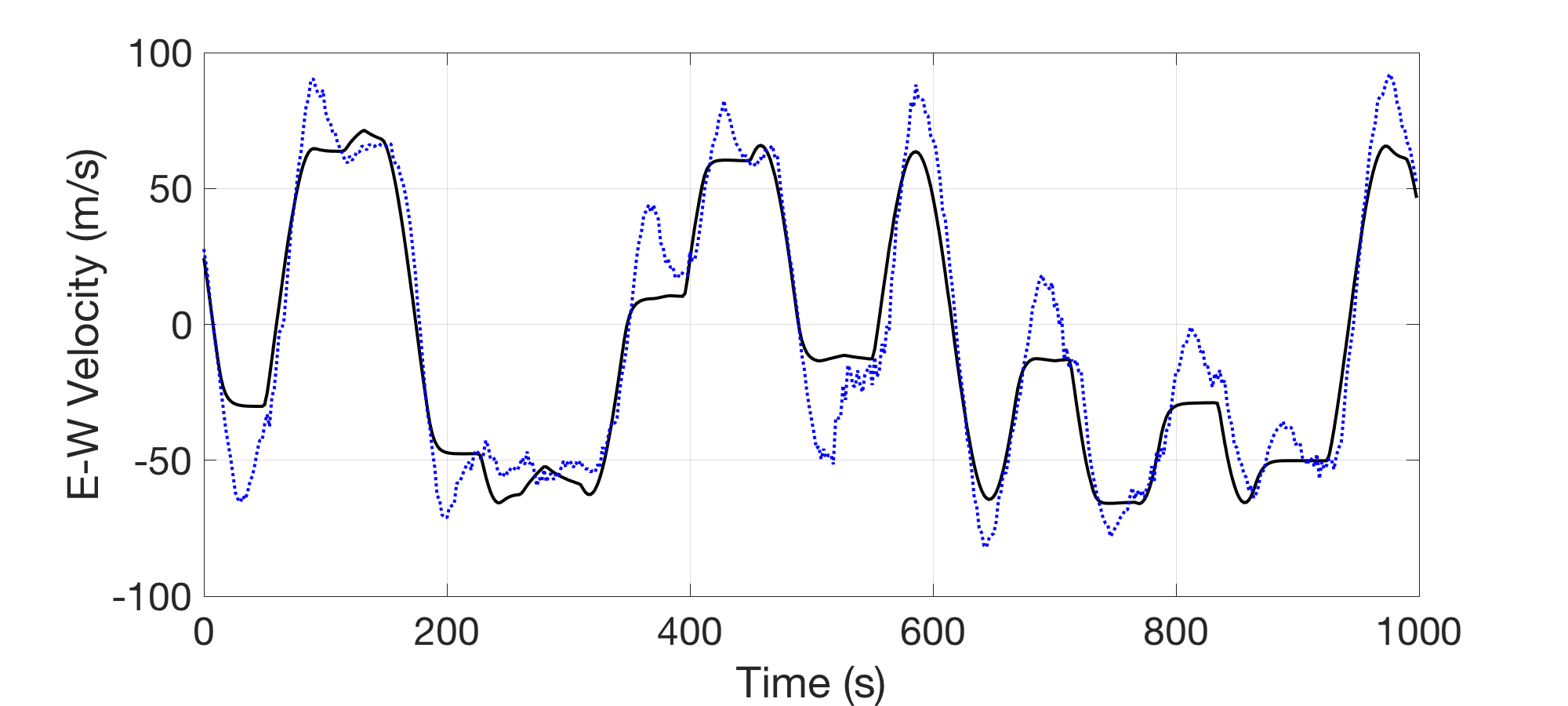}}
	\subfigure[PRSSM]{
		\includegraphics[width=0.45\columnwidth]{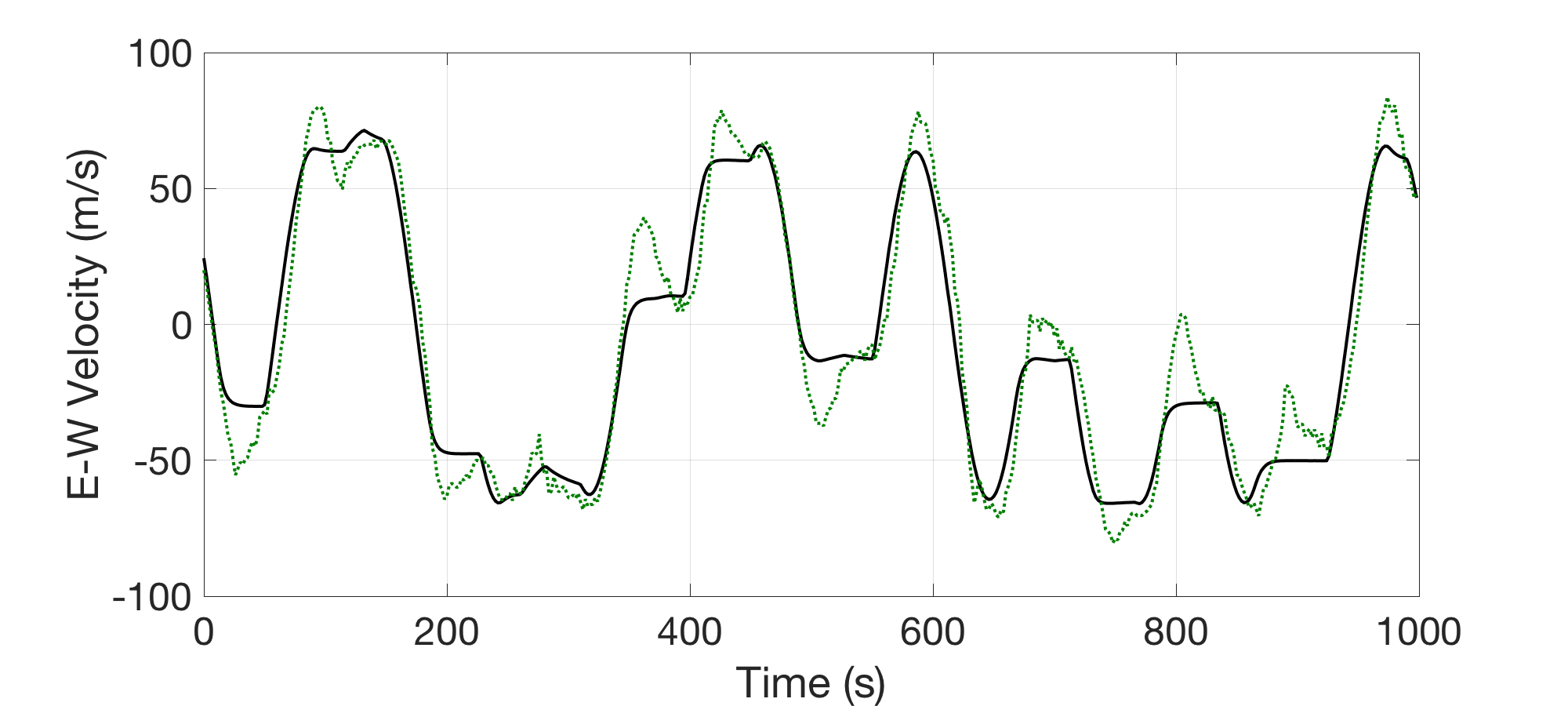}}
	\subfigure[H-GPSSM]{
		\includegraphics[width=0.45\columnwidth]{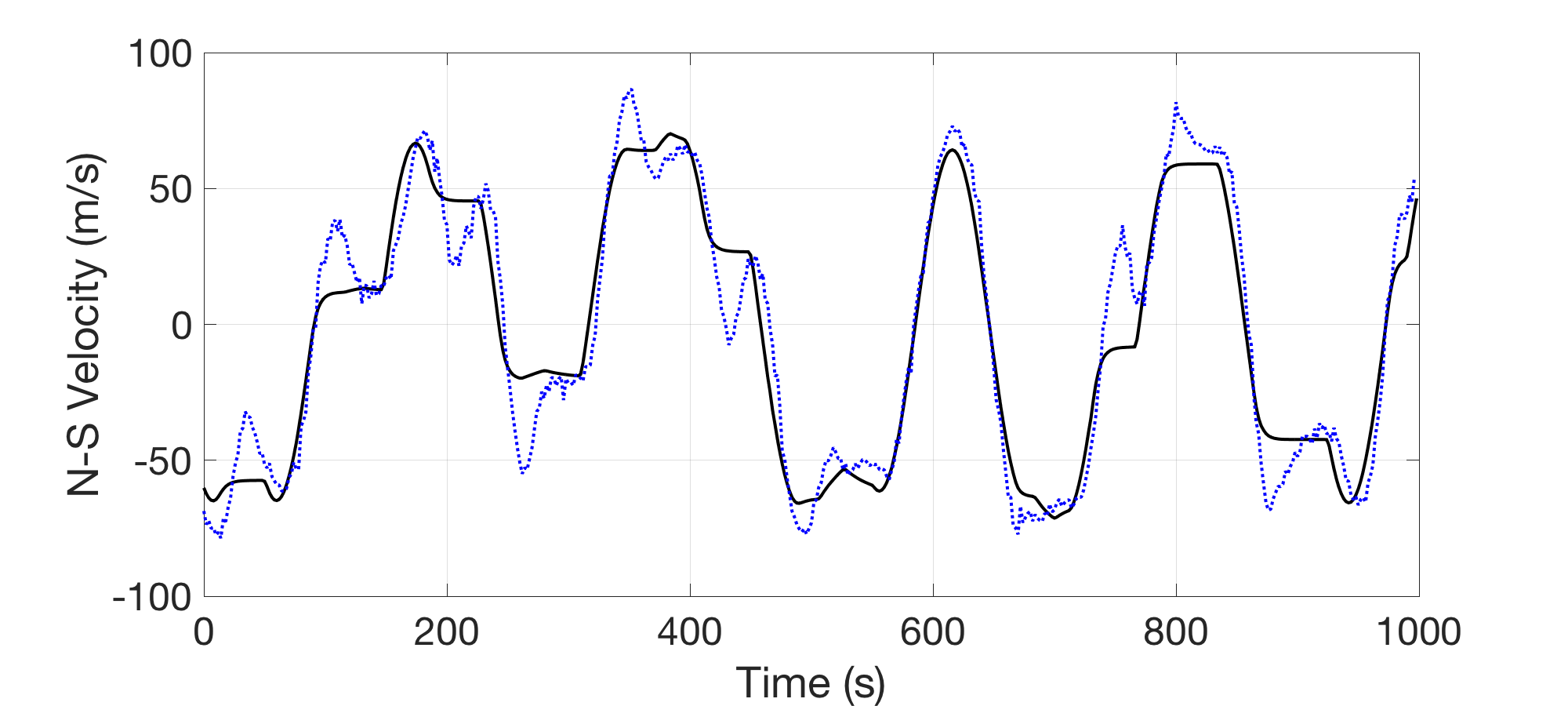}}
	\subfigure[PRSSM]{
		\includegraphics[width=0.45\columnwidth]{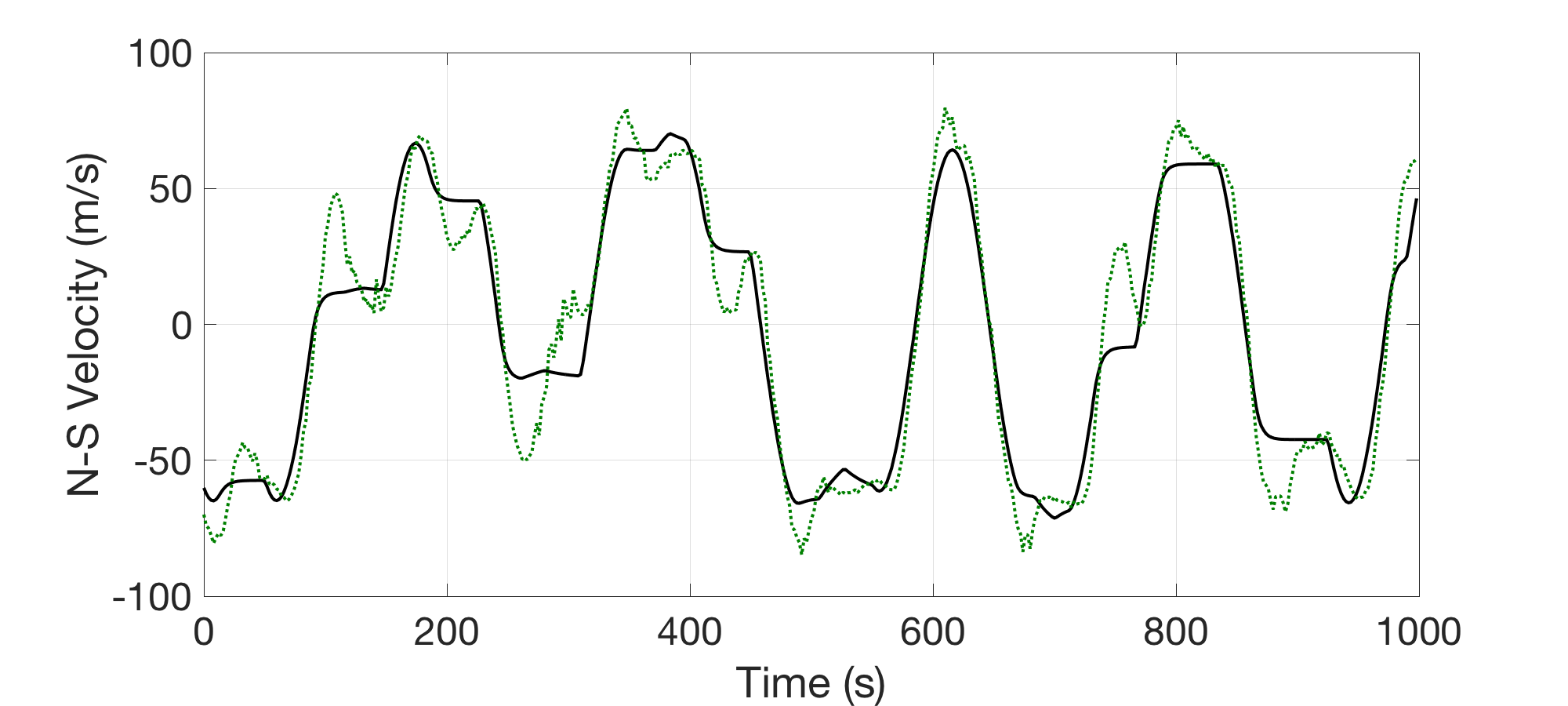}}
	\caption{ATC tracking results by using H-GPSSM (blue) and PRSSM (green) and particle filter. Ground truth and estimated states are colored by black.}
	\label{fig:ATC2}
\end{figure}

\end{document}